\documentclass{article}

\PassOptionsToPackage{numbers, compress}{natbib}
\usepackage[preprint]{neurips_2025}

\usepackage[utf8]{inputenc} % allow utf-8 input
\usepackage[T1]{fontenc}    % use 8-bit T1 fonts
\usepackage{hyperref}       % hyperlinks
\usepackage{url}            % simple URL typesetting
\usepackage{booktabs}       % professional-quality tables
\usepackage{amsfonts}       % blackboard math symbols
\usepackage{nicefrac}       % compact symbols for 1/2, etc.
\usepackage{microtype}      % microtypography
\usepackage{xcolor}         % colors

\usepackage{graphicx,graphbox}
\usepackage{bbm}
\usepackage{dsfont}
\usepackage{multirow}

\usepackage{amsmath}
\usepackage{enumitem}  % Add this in the preamble

\usepackage{wrapfig}
\usepackage[ruled]{algorithm2e}
\usepackage{tikz}
\usetikzlibrary{arrows.meta, positioning}
\usetikzlibrary{shapes,arrows,chains}
\usetikzlibrary[calc]
\usepackage{adjustbox}
\tikzstyle{process} = [rectangle, draw, text width=5em, text centered]
\tikzstyle{line} = [draw, -latex']
\tikzstyle{input} = [trapezium, draw, minimum width=2cm, text width=3em, trapezium left angle=120, trapezium right angle=60, text centered]
\tikzset{>=stealth}

\title{Learning Genomic Structure from \(k \)-mers %  Reads

}

\author{%
  Filip Thor\thanks{Corresponding author} \\
  Division of Scientific Computing, \\
  Department of Information Technology\\
  Science for Life Laboratory\\
  Uppsala University\\
  Uppsala SE-752 37, Sweden \\
  \texttt{filip.thor@it.uu.se} \\
  \And
    Carl Nettelblad \\
  Division of Scientific Computing, \\
  Department of Information Technology\\
  Science for Life Laboratory\\
  Uppsala University\\
  Uppsala SE-752 37, Sweden \\
  \texttt{carl.nettelblad@it.uu.se} \\
}

\begin{document}

\maketitle

\begin{abstract}
Sequencing a genome to determine an individual's DNA produces an enormous number of short nucleotide subsequences known as reads, which must be reassembled to reconstruct the full genome. 
We present a method for analyzing this type of data using contrastive learning, in which an encoder model is trained to produce embeddings that cluster together sequences from the same genomic region. The sequential nature of genomic regions is preserved in the form of trajectories through this embedding space.
Trained solely to reflect the structure of the genome, the resulting model provides a general representation of $k$-mer sequences, suitable for a range of downstream tasks involving read data. 
We apply our framework to learn the structure of the \textit{E.\ coli} genome, and demonstrate its use in simulated ancient DNA (aDNA) read mapping and identification of structural variations. Furthermore, we illustrate the potential of using this type of model for metagenomic species identification.
We show how incorporating a domain-specific noise model can enhance embedding robustness, and how a supervised contrastive learning setting can be adopted when a linear reference genome is available, by introducing a distance thresholding parameter \(\Gamma\). 
The model can also be trained fully self-supervised on read data, enabling analysis without the need to construct a full genome assembly using specialized algorithms. 
Small prediction heads based on a pre-trained embedding are shown to perform on par with BWA-aln, the current gold standard approach for aDNA mapping, in terms of accuracy and runtime for short genomes. 
Given the method's favorable scaling properties with respect to total genome size, inference using our approach is highly promising for metagenomic applications and for mapping to genomes comparable in size to the human genome. In both scenarios, the flexibility of not being tied to a single linear reference can provide a more nuanced analysis of input data.

%DNA short read mapping, which means finding where short sub-sequences fits within a genome. Using contrastive learning, we construct embeddings that reflect the e of the E-coli genome, while being robust to data noise. 
%Using the embeddings, we perform mapping of ancient DNA reads, which is difficult both because of a reduced read length and noisier data. 
%Also, we introduce a general way of doing regression on a large target domain. Instead of viewing it as a regression task, we train a model to predict the binary representation of the position, with better computational scaling than a equisized binning approach. We also show that a tiny GPT model is a more accurate predictor than an MLP for this scenario. 
%[Not yet written]

%The abstract paragraph should be indented \nicefrac{1}{2}~inch (3~picas) on
%both the left- and right-hand margins. Use 10~point type, with a vertical
%spacing (leading) of 11~points.  The word \textbf{Abstract} must be centered,
%bold, and in point size 12. Two line spaces precede the abstract. The abstract
%must be limited to one paragraph.

\end{abstract}

\section{Introduction}

Modern next-generation sequencing (NGS) platforms generate massive numbers of short subsequences of DNA referred to as \textit{reads}, which consist of sequences of nucleotide bases \(A, T, C\) and \(G\) \cite{Metzker2010}. % but lacking information about where they came from in the genome. %The data we consider in this work are such short sequences of nucleotide bases \(A,T,C\) and \(G\). 
Reads contain no inherent positional information. Therefore, reconstructing the full genome of an individual from reads requires reassembling them, which is typically done by aligning them to the most similar region in a reference genome -- a process known as \textit{read mapping} or \textit{read alignment }\citep{Alser2021}. Performing a unique read alignment proves challenging when there are repetitive or highly similar segments in the genome. When reads are shorter, ambiguities become more prevalent \citep{Logsdon2020}.
One of the leading short read sequencing platforms, Illumina, produces reads that can vary from 30 to 500 base pairs (bp) \citep{Goodwin2016}, but recent developments in DNA sequencing have made long read sequencing more reliable and available, producing reads of lengths 10k to 100k bp \cite{Marx2023}. This greatly facilitates alignment, as most regions are unique at those length scales.
In this work, we analyze sequences 30bp in length, which would represent read lengths that can occur in ancient DNA (aDNA) sequencing \cite{Bettisworth2025-ug}. aDNA is characterized by specific damage patterns, such as cytosine deamination that increases individual nucleotide errors, and also suffers from severe fragmentation \cite{briggs}. 
Fragmentation of genetic material prevents the recovery of longer reads, regardless of advancements in sequencing technology.
Thus, robust methods for analyzing ultra-short read data are still needed for us to learn about our past through studying genetics.

\paragraph{Our contribution.} We introduce a framework we have named Contrastive Read Network (CReadNet). Inspired by recent developments in contrastive representation learning, we demonstrate how an encoder model can be trained to map short \textit{k-mers} (a sequence of $k$ nucleotides) to a general embedding reflecting the genomic structure. 
This is done by extracting two $k$-mers with a small genetic distance (measured in bps), minimizing their distance in the embedding space, and maximizing the distance to other \(k\)-mers. Due to transitivity, this yields a continuous embedding that reflects the ordering of the \(k\)-mers within the genome. 
Depending on the data available, we frame the problem in either a supervised or self-supervised contrastive learning setting. %We focus on the setting of read mapping, where access to a reference genome is assumed, and positive \(k\)-mers can be created by simply sampling two starting coordinates in close proximity. If we have a non-model species with no reference and only read data available, \(k\)-mers can be generated by drawing subsequences from the same read and processing those.
This training is task-agnostic, producing a general representation that can be used for a diverse set of tasks. We demonstrate how a pre-trained encoder model can be used to identify both structural variation in a genome and discriminate between disjoint sequences. The latter has a potential for applications in metagenomics. In our experiments, we focus on aDNA read mapping. For this task, we show that a model predicting the bitwise representation of the read position outperforms a regression model of equal size. We also show that using a small GPT to sequentially predict bits improves the accuracy further. %  We train 
%perform read mapping, and we introduce a bit-predicting GPT head that outperforms a simple MLP prediction head when trained either as a regression or a classification problem. We also demonstrate that the pre-trained embedding can be used for other applications, such as finding inversions, and identifying disjoint sequences. 

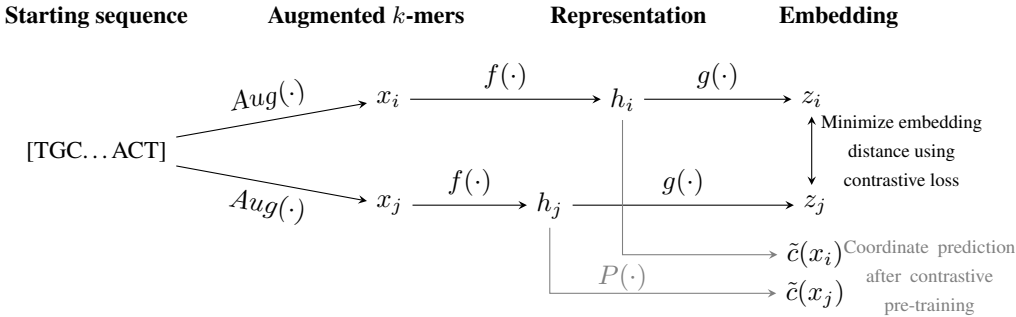
\begin{figure}[htb]
\resizebox{\linewidth}{!}{  
\begin{adjustbox}{width=\linewidth, trim=0 0 0 0, clip}

\begin{tikzpicture}[node distance=2cm, every node/.style={align=center}, scale=0.50]

% Nodes
\node (read) at (0, 0) {\small [TGC\dots ACT]};

\node (x1) [above right=.2cm and 2.5cm of read] {\(x_i\)};
\node (x2) [below right=.2cm and 2.5cm of read] {\(x_j\)};

\node (r1) [right=2.5cm of x1] {\(h_i\)};
\node (r2) [right=1.5cm of x2] {\(h_j\)};

\node (e1) [anchor = center, right=5cm of x1] {\(z_i\)};
\node (e2) [anchor = center, right=5cm of x2] {\(z_j\)};

\node (predict1) [below =0.1cm of e2] {\(\tilde{c}(x_i)\)};
\node (predict2) [below=0.6cm of e2] {\(\tilde{c}(x_j)\)};
\node (temp_pred) [below right=1.5cm and 1.5cm of r2] {};
\node (temp_pred2) [below=2.5cm  of r2] {};

% Arrows
\draw[->] (read) -- node[midway, above, sloped] {\(Aug(\cdot)\)} (x1);
\draw[->] (read) -- node[midway, below, sloped] {\(Aug(\cdot)\)} (x2);

\node (test) at ([yshift=3cm]read.north) {\small \textbf{Starting sequence}};

\node (augmentation) [ right=0.8cm of test] {\small \textbf{Augmented \(k\)-mers}};

\node (projection) [right=4.5cm of test] {\textbf{\small Representation}};
\node (embedding) [right=7.5cm of test] {\small \textbf{Embedding}};

\draw[->] (x1) -- node[midway, above] {\(f(\cdot)\)} (r1);
\draw[->] (x2) -- node[midway, above] {\(f(\cdot)\)} (r2);

\draw[->] (r1) -- node[midway, above] {\(g(\cdot)\)} (e1);
\draw[->] (r2) --node[midway, above] {\(g(\cdot)\)}  (e2);

%\draw[<->] (e1) -- node[right, midway] {\scriptsize Minimize embedding \\ \scriptsize distance   using  \\ \scriptsize contrastive loss  } (e2);
\draw[<->] (e1) -- ++(0,-2.25cm) node[right, midway] {\scriptsize Minimize embedding \\ \scriptsize distance using \\ \scriptsize contrastive loss};

\node at ($(predict1)!0.5!(predict1)$) [xshift = 1.5cm,yshift = -0.3cm, text width =2.5cm] {\scriptsize \color{gray} Coordinate prediction after contrastive pre-training};
\draw[gray, ->] (r2) |- node[below, midway] {}(predict2);
\draw[gray, ->] (r1) |- node[below, midway] {\(P(\cdot)\)}(predict1);
\end{tikzpicture}%
\end{adjustbox}
}
\vspace{-2cm} % Adjust this value as needed

\caption{Given a starting sequence, we draw two \(k\)-mers, \(Aug(\cdot)\) adds aDNA noise, resulting in pairwise positive \(k\)-mers \(x_i\) and \(x_j\).
A convolutional encoder model \(f\) and a one-layer projection layer \(g\) are trained to minimize the embedding distances between \(z_i\) and \(z_j\). The intermediate representation \(h_i\) is used by a prediction head \(P\) when predicting \(k\)-mer coordinates \(c_i\).
%A neural network model \(f\circ g(\cdot)\), where \(f\) is a convolutional encoder model, and \(g\) is a one layer projection the is trained to minimize the embedding distances \(z_i\) and \(z_j\).
%which act as positives in the contrastive loss.
%\(Aug(\cdot)\) denotes the augmentation module, which consists of the sampling of \(k\)-mers and application of domain-specific noise.  \(f(\cdot)\) denotes the encoder network mapping \(k\)-mers \(x_i\) to representations \(h_i\), \(g(\cdot)\) a single layer projection layer producing the embeddings \(z_i\), and the position prediction network by \(P(\cdot)\) used in outputting predicted coordinate locations \(\tilde{c}(x_i)\) in the read mapping task. 
      }
      \label{fig:pipeline}
  \end{figure}

The framework is illustrated in Figure \ref{fig:pipeline}. Our main contributions can be summarized in three points:
%\begin{itemize}
\begin{itemize}[leftmargin=*, itemsep=1pt, labelsep=0.5em]
\item We apply contrastive learning to train a neural network that learns a continuous representation of a genome based on short \(k\)-mers, or sequences of base pairs.
%\item We use a domain-adapted augmentation scheme, making the model robust to both read complements and to aDNA deamination patterns.
\item We demonstrate how a domain-adapted augmentation scheme makes the model robust to intricacies found specifically in genetic read data. % both read complemen ts and to aDNA deamination patterns.
\item We frame the read mapping problem as a classification problem on a bitwise representation of the coordinate, improving over regression, and scaling better than predicting equidistant bins.  
\end{itemize}

\paragraph{Related work.}

There has been some work in genome alignment using deep learning approaches. BetaAlign \citep{betaalign} uses transformers to do multi-sequence alignment, matching groups of 10 reads with each other. Similarly, \citet{Lall2023} does pairwise read alignment using reinforcement learning and focuses on developing models tailored for edge devices.
\citet{delux} uses unsupervised learning to cluster long reads (\(\sim\)10kbp).
\citet{google_seq} uses a convolution-based NN to do species identification from short reads.
\citet{read_aln} does short read genome alignment. In contrast to our work, their model treat alignment as a single multi-class classification problem, binning the reference genome into several classes, and predicting the approximate position of a given read. Our work demonstrates how a pre-trained meaningful general representation of the DNA structure can be used for several applications, not only read mapping, and we show how to target the difficulties arising when analyzing aDNA reads specifically.

\section{Methodology}

The core principle in contrastive learning is simple: each sample in a batch gets designated \textit{positive} and \textit{negative} samples, and a neural network is trained using a loss that attracts the positives, and repulses the negatives \cite{facenet,simCLR,npair}. In this work, we adapt this methodology and apply it to genetic read data. 
We aim to create an embedding of \(k\)-mers that reflects the genomic structure, which can be used for downstream tasks involving short-read data. 
%DNA strings have an inherently linear and sequential structure, and given error-free subsequences of length \(k\) and that the genome we consider has no repeats longer than \(k\), we should be able to recreate the genome structure by placing the \(k\)-mers along a threaded line in the embedding space, resembling a string of beads. 
%However, with aDNA reads, we have very short and error-prone reads, so we need a method that is robust to the types of noise we can expect, and there is no hope to fully resolve all sections of the genome. Rather, we would expect that there could be loops, and perhaps stretches of interlaced reads stemming from different parts of the genome -- turning the beaded string into a tangled ball of yarn. Still, we believe that learning a continuous embedding would be a good base for downstream prediction tasks.
%In this section, we discuss both the deep learning methodology we have developed to learn the genomic structure from short sequences, and the properties inherent to genetic read data that needs to be considered when training a neural network.
Following the workflow in Figure \ref{fig:pipeline}, we start by introducing the data and the augmentation scheme, followed by our customized loss function and the neural network models used.
\paragraph{ \(\mathbf{\textit{k}}\)-mers, Directionality, and Read Complements.}
We consider \(k\)-mers, substrings \(k\) nucleotide bases in length extracted from some longer sequence. For example, the sequence [TGCGTGG] has the following 3-, 4- and 5-mers
%\[[\text{ATGTAC}] \to\begin{cases}
%\text{3-mers: \{[ATG],[TGT],[GTA],[TAC]\}},\\ 
%\text{4-mers: \{[ATGT],[TGTA],[GTAC]\}},\\
%\text{5-mers: \{[ATGTA], [TGTAC]\}}.\end{cases}\]
\[
[\text{TGCGTGG}] \to \begin{cases}
\text{3-mers: \{[TGC], [GCG], [CGT], [GTG], [TGG]\}},\\ 
\text{4-mers: \{[TGCG], [GCGT], [CGTG], [GTGG]\}},\\
\text{5-mers: \{[TGCGT], [GCGTG], [CGTGG]\}}.
\end{cases}
\]

\begin{figure}[ht]
    \centering
    \includegraphics[width=0.8\linewidth]{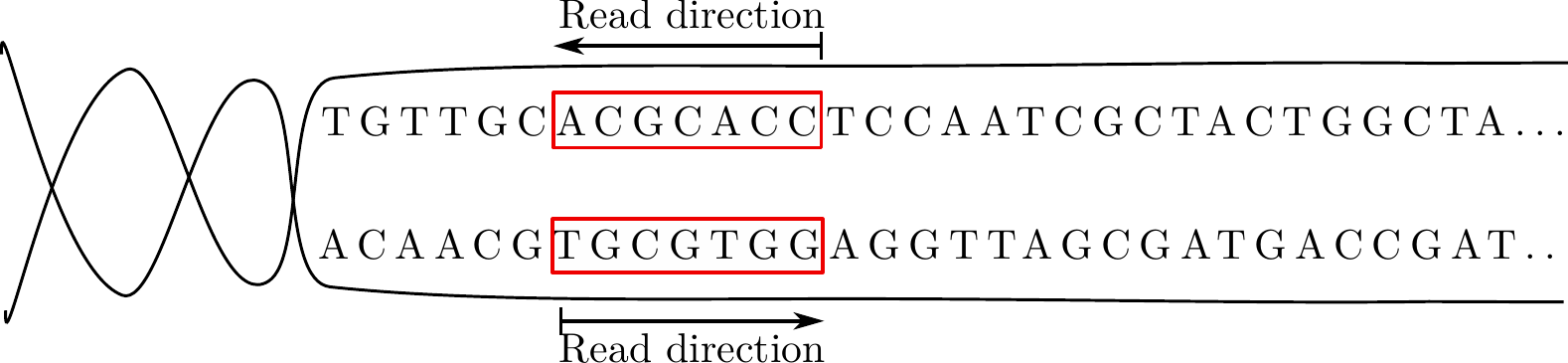}

    \caption{ Reads from the same position in the genome can be expressed differently based on which strand the sequence originates from, [TGCGTGG] and [CCACGCA] have the same bp coordinate.
    %Illustration of the read direction in Illumina short read sequencing. 
    %The direction of reads is unknown, and thus we need the encoder to be robust to the complementary.
    %We want nucleotide sequences and their reversed complementary sequence to be identified as the same. 
    }
    \label{fig:reads}
\end{figure}

%Predicting the exact position of a short read is difficult, and we instead aim to predict an approximate position, after which local alignment can be performed. This refinement step is done by sliding the read across a small window (e.g., 2500 bps) near the predicted location, and finding the best match. This short-window local alignment can be efficiently parallelized on a GPU, while attempting to do this over the full genome is computationally infeasible.
%This means that accurately predicting the exact correct position is hard, and we relax the problem to be content if we are "sufficiently close", and predict an approximate position. After that, we can apply a local alignment method. Finding the best match of a 30bp read in a 2500bp window can be done by naively sliding the read along the window, and recording the best match. This can be done in parallel on a GPU, while attempting do this over the full genome is infeasible.
DNA sequences are double-stranded, and in short-read genome sequencing, one does not know which strand of the double helix the sequence comes from.
%At creation, we do not know the direction of the reads,  it could have been read from either strand in the double helix. 
This means that a read can be expressed in terms of its \textit{reverse complement}, i.e., in reversed order and with the bases changed as \(A \leftrightarrow T\) and \( C \leftrightarrow G \) when compared to the reference genome, which describes just one of the strands.
Figure \ref{fig:reads} illustrates an excerpt of the double helix, and shows that reads from the same genomic position can be expressed differently depending on which strand the read is taken from. To handle real short-read data, the model must recognize a \(k\)-mer and its reverse complement as similar.
%Thus, we need a model that can recognize and align both the forward read, and it's complement.
%Figure \ref{fig:reads} shows an illustration this.

We use a one-hot encoding of the input \(k\)-mer as input to our encoder model. By choosing the base ordering \(A,C,G,T\), the reverse complement is obtained by simply flipping both axes in one-hot encoded data, as shown in Equation \eqref{eq:comp}. 
\begin{equation} \label{eq:comp}\text{RevComp}(\text{[TGCGTGG]}) = \text{RevComp} \left(\begin{bmatrix}  
0 & 0 &0 &1 \\
0 &0 &1&0\\
0& 1& 0& 0
\\ 0 & 0 & 1 & 0
\\ 0 & 0 & 0 & 1
\\ 0 & 0 & 1 & 0
\\ 0 & 0 & 1 & 0
\end{bmatrix} \right)  = \begin{bmatrix}  
0 & 1 &0 &0 \\
0 &1 &0&0\\
1& 0& 0& 0
\\ 0 & 1 & 0 & 0
\\ 0 & 0 & 1 & 0
\\ 0 & 1 & 0 & 0
\\ 1 & 0 & 0 & 0
\end{bmatrix} = \text{[CCACGCA]}
\end{equation}

\paragraph{Sequencing Errors and aDNA Damage.}
In NGS, many types of errors can contaminate the produced reads. The most common error types are substitutions, where one base is substituted by another, and indels, where a base can either be deleted or inserted between two bases. Here, we try to mimic Illumina sequencing, where substitutions are the most common, and have a prevalence of \(0.1\)--\(1\%\) \cite{error_rate}.
%This type of error is inherent to the sequencing tool used, and is a measurement error that can occur in all reads. 
We want to apply our method to aDNA, which has additional characteristic damage patterns. 
Aged DNA samples have drastically reduced fragment length, shortening the read length.
They also suffer from deamination damage, a chemical process that breaks down the cytosine base. This is expressed as a higher error rate for the \(C \to T\) and \(G \to A\) substitutions close to the ends of the reads \cite{briggs}. We want the model to be robust to both the read complement and the types of noise we expect to see in real aDNA data. 

\paragraph{Augmentation Scheme.} 
In contrastive learning, we train a model to minimize distances between \textit{similar} samples. The augmentation scheme applied to the positives is important, since we then implicitly state that two augmented versions of a sample are still supposed to be considered to be \textit{similar}. Ideally, the augmentation should cover the kinds of deviations from the non-augmented training data that can be expected during inference.
We want to train the model to attract two \(k\)-mers if they \textbf{1:} have a small distance in the genome, \textbf{2:} are reverse complements of each other, and \textbf{3:} are noisy versions of the same \(k\)-mer. An embedding that satisfies these three conditions captures the genomic structure faithfully.  
Combining these three attributes, we use the following augmentation scheme \(Aug(\cdot)\). Given a starting sequence of length \(L\), and a maximum offset between positive \(k\)-mers \(d\), %we generate a pair of \(k\)-mers. If a reference genome is available,
we sample the starting coordinate of the first \(k\)-mer \(x_i\)  as \(c_i \sim \mathcal{U}(0,L-k-d)\), and the coordinate of its positive \(x_j\) as \(c_j \sim \mathcal{U}(c_i,c_i + d)\). With a probability of 0.5, we apply the reverse complement transformation separately for all \(k\)-mers, and add a flat \(1\%\) substitution rate in all positions. We restrict ourselves to considering \(k\)-mers of length 30 to be able to handle the short read lengths in aDNA, and to model the deamination noise, we increase the substitution rate for  \(C \rightarrow T\) in the first 10 bases and \( G \rightarrow A \) in the last 10 bases to \(10\%\).
%If we have a reference genome available, the starting sequence for all samples is the reference genome. 
The starting sequence will be the reference genome if available, otherwise, the pair of positive $k$-mers will be drawn from the same read of larger total length.%, the starting sequences will be reads, ensuring that the positive from the same read as the starting sequence.

%The approach we take is to first train an encoder model to produce a general embedding, trained using contrastive learning inspired by the simCLR framework \cite{simclr} and its supervised counterpart \cite{supcon}. 
%Read mapping, or the position prediction step, is later done using a prediction head model, which takes the output of the pretrained model. %  a methodology that should generalize to other similar regression tasks. %These two main components are discussed in this section.

\paragraph{Coordinate-Thresholded Contrastive Loss.}

%embedding error-free reads longer than \(k\) this structure could be resolved fully.
%The reads could be placed along a threaded line in the embedding space, resembling a string of beads.
%However, with aDNA reads, we have very short and error-prone reads, so we need a method that is robust to the types of noise we can expect, and there is no hope to fully resolve all sections of the genome. Rather, we would expect that there could be loops, and perhaps stretches of interlaced reads stemming from different parts of the genome -- turning the beaded string into a tangled ball of yarn. Still, we believe that learning a continuous embedding would be a good base for downstream prediction tasks.

%so there is no hope of fully resolving all sections of the genome, and we need a method that is robust to the types of errors we can expect.

%Genomic data has a strictly linear structure, it is simply a long, 1D sequence of nucleotide bases. Given a read length \(L\), if there is no repeated sequences longer than \(L\), we would expect to be able to place the reads in a straight line. 
%However, if there are repeats longer than \(L\), we can no longer have a model that can would be able to place them along a line. 

%We ideally want to create an embedding that places pairs of reads that have similar coordinates close to each other. With a transitive property, we would then expect that this would create an embedding resembling a string of beads in d-dimensions.    

We train the model using an altered version of the simCLR loss, which can be used in a  supervised as well as a purely self-supervised contrastive learning setting. 
%To create the embedding, we employ an encoder network \(f\)and a small projection head \(g\), and an altered version of the supervised regime of the contrastive learning framework.  
%The framework has three main components; the augmentation scheme \(A(\cdot)\) which augments and produces two views of each read, the encoder model \(f(\cdot)\) mapping the reads \(x_i\) to a representation \(h_i\), and a one layer projection model \(g(\cdot)\) mapping the representation to the embedding \(z_i\). 
We use the following definition for the loss function, following the notation in \citep{supcon}
\begin{equation}
    \mathcal{L} = \sum_{i\in I} \frac{-1}{\vert P(i)\vert}    \sum_{p\in P(i)}d_{i,p} \log \frac{\exp (z_i \cdot z_p / \tau ) }{ \sum_{a \in A(i) }\exp (z_i \cdot z_n / \tau)}, 
\end{equation}
where \(z_i \) are the \(k\)-mer embeddings produced by the model, \(\tau\) is a temperature parameter, \(I = \{1,\dots,2N\}\) the set of all indices in a batch, and \(A(i) = \{I\backslash i\} \)  and  \(P(i)\)  the set of negatives and positives of \(k\)-mer  \(i\), respectively. %. \(i\), and \(P(i)\) the set of the positive of \(k\)-mer  \(i\).
%Most contrastive learning literature focuses on computer vision and image classification. In that case, the set of positives is taken as the samples in the batch that share the same label as \(i\). 
In self-supervised learning, the positive is just the positive pair from the augmentation, \(P(i)= j\). When we have access to a reference genome and thus know the coordinates \(c_i\) used in training, we can extend the set of positives to include all nearby \(k\)-mers by introducing a thresholding parameter \(\Gamma\), and define the positives as   
%Since the label information in our case is coordinate position along the genome, we do not have the same label structure with fixed groups of samples sharing the same label.
%Instead, we consider samples to be positives of one another if they are sufficiently close, and for a thresholding value \(\Gamma \), we define the set of positive samples as
\begin{align}P(i) =\begin{cases} \{ p\in \mathcal{I} \mid \lvert c_i - c_p \rvert \leq \Gamma \}, &\text{if supervised}\\
j, &\text{if self-supervised}.
\end{cases}\end{align} 
This choice has two main benefits. We increase the number of positives, which has been shown to positively impact performance \citep{supcon}. When using large batches and large augmentation offsets \(d\), it will be common for a supposed negative to be closer to the anchor than the positive. This means that the gradients would have repulsion terms from samples that are genomically closer than the assigned positive. \(\Gamma\)-thresholding makes those samples positives, instead.
If the augmentation offset \(d<\Gamma\), we are guaranteed to always have at least one positive per sample. In the worst case the supervised loss collapses to the self-supervised formulation.
%Since we create two views from each sample, ensuring that each sample has at least one positive. 
We also introduce a distance scaling in the thresholded loss, which weights the loss contributions by the distance between the anchor and each positive, defined as \(d_{i,p} = \frac{\vert  c_i - c_p\vert}{\Gamma} \), which in the self-supervised setting is just set to 1.

%Here, we have assumed access to the reference genome during training of the embedding, which is the case for the standard read mapping setups. However, the framework can be applied in non-model species which does not yet have a reference genome, if we have longer reads than the k-mer size we consider. By simply removing the distance scaling \(d_{i,p}\), and defining positives as different k-mers from the same read, we can train the encoder self-supervised, instead. 

\paragraph{Encoder Model Architecture.}

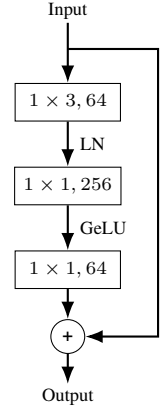
\begin{wrapfigure}{r}{0.15\textwidth}  % 'r' for right, 'l' for left
  \centering
  \vspace{-1em}  % optional: adjust vertical placement
  
\begin{tikzpicture}[
    box/.style={draw, minimum width=1.4cm, minimum height=0.5cm, align=center, font=\scriptsize},
    arrow/.style={-{Latex}, thick},
    skip/.style={-{Latex}, thick},
    sum/.style={circle, draw, minimum size=4.5mm, inner sep=0pt, font=\scriptsize\bfseries},
    node distance=0.4cm
]

% Nodes
\node (input) at (0,0.2cm) {};
\node[box, below=0.8cm of input] (conv1) {\(1\times 3, 64\)};
\draw[arrow] (input) -- (conv1) node[pos=-0.2, ] {\scriptsize Input };

\node[box, below=0.6cm of conv1] (ln) {\(1\times 1, 256\)};
\draw[arrow] (conv1) -- (ln) node[midway, right=1pt] {\scriptsize LN};

\node[box, below=0.6cm of ln] (gelu) {\(1\times 1, 64\)};
\draw[arrow] (ln) -- (gelu) node[midway, right=1pt] {\scriptsize GeLU};

\node[sum, below=of gelu] (sum) {+};
\draw[arrow] (gelu) -- (sum);

\node (output) [below=of sum] {};
\draw[arrow] (sum) -- (output) node[pos=0.05cm, ] {\scriptsize Output};

% Midpoint of input to conv1 arrow
\coordinate (midskip) at ($(input)!0.4!(conv1)$);
\coordinate (skipout) at ($(sum)+(1.2,0)$);
\draw[arrow] (midskip) -- ++(1.2,0) -- (skipout) -- (sum);

\end{tikzpicture}

  \caption{\\Residual block layout. %with 4x inverted bottleneck, layernorm, and GeLU activation .
  }
  \label{fig:residual}
  \vspace{-0.5cm}  % reduce space after figure

\end{wrapfigure}

We use a ConvNet as the encoder network since we want a translationally invariant model that captures local patterns within the \(k\)-mers. We adopt a similar residual block and overall structure as the ConvNeXt model by \citet{Liu_2022_CVPR}, with some alterations tailored toward short DNA sequences.
Since the reads are very short, we use stride 1 in the input stem, and use regular rather than depthwise convolutions. 
The residual block structure is shown in Figure \ref{fig:residual}. After each stage of residual blocks, the input is downsampled using an average pooling with stride 2 and kernel size 2,  followed by a convolution with kernel size 3, doubling the channels.

We use the same compute ratio as ConvNeXt with four compute stages, and we scale the model as:
%\begin{itemize} 
\begin{itemize}[leftmargin=*, itemsep=1pt, labelsep=0.5em]

    \item CReadNet-T: Channels \( = (64, 128, 256, 512), \) 
      Blocks\( =  (3, 3, 9, 3 ) \), \(   16.8 \)M params,
    \item CReadNet-S: Channels \( = (64, 128, 256, 512),\) Blocks  
      \( =  (3, 3, 27, 3 ) \), \(   29.8 \)M params,
    \item CReadNet-B: Channels \(
    = (96, 192, 384, 768),\)  Blocks\(  =  (3, 3, 27, 3 ) \), \( 66.5 \)M params.  \end{itemize} 

%is summarized in Table \ref{tab:}
%Since we use contrastive learning, and a very wide skinny (thin and long) input space, we want to be able to use as large per-device batch size as possible to improve the quality of the positive and negatives. Due to this, we use an architecture with less residual block, and stick with 3 for each stage.
%The first residual block consists of a depthwise conv1D layer with kernel size \(k=5\), followed by two conv1D with \(k = 1\), and 192 channels.
%Layer normalization is applied after the first layer, and GeLU after the second.  After each series of residual block, the network downsamples using a conv1D with   \(k=2, s=2 \), and doubles the channels.  After the residualblocks Global Average Pooling is applied, the output of which is the 

\iffalse
\begin{table}[htb]%{}{0.6\linewidth}
\caption{Encoder model architecture} \label{wrap-tab:1}
 \begin{tabular}{lll}
    \toprule
    %\multicolumn{2}{l}{Encoder Architecture %(1D ConvNeXt)
    %}                   
    %\cmidrule(r){1-2}
%    Name     & Description   Params \\
 %   \midrule
    %Input Stem & \(k= 5, C=64, s = 1 \)\\
    
    %Channels & \(C = (64, 128, 256, 512)\) &      \\
    %Blocks     & \(B =  (3, 3, 9, 3 ) \)&      \\
 %   Downsampling     & \(k = 2, s = 2\)       &   \\
    CReadNet-T & \(C = (64, 128, 256, 512), \) 
     \( B =  (3, 3, 9, 3 ) \)  \\
    CReadNet-S & \(C = (64, 128, 256, 512),\)  
      \(B =  (3, 3, 27, 3 ) \) \\
    CReadNet-B & \(C = (96, 192, 384, 768),\) \( B =  (3, 3, 27, 3 ) \)  \\ 
    \bottomrule
  \end{tabular}
\end{table} 

\fi
The encoder model ends with global average pooling and a dense layer with the same number of units as the output channels pre-pooling, which produces the representation \(h_i\). A last linear layer then outputs the 256-dimensional embeddings \(z_i\). Both  \(z_i\) and \(h_i\) are \(L_2\) normalized. % in which the contrastive loss is applied.

\paragraph{Position Prediction.}
\label{sec:pos_pred}
The main application we target is read alignment, which means finding the position of a read along a reference genome. Here, we define the coordinate of a \(k\)-mer as the position of its first nucleotide. 
%One task we focus on in this work is the prediction of the position \(c(x)\) along a reference genome given a short DNA read \(x\in \mathcal{X}\), where the position is defined as the coordinate of the first base in the read. Genomes vary in size but are typically very large. For example, the  \textit{E.coli} genome is 4.5 million base pairs long, while the human genome spans 3 billion bases.
Analogous to classification heads used in image classification, we employ a smaller network \(P(\cdot)\) to perform the mapping task. This model is trained using the representation \(h_i\), with the encoder model weights frozen. We compare three different ways of doing this, all starting with a 3-layer MLP with 2,048 units, using layer normalization before, and SiLU activation after, in each layer. The first approach is using a regression model, where the output of the MLP is the predicted coordinate \(\tilde{c}(x)\), and the model is trained using the MSE loss, \(L_{MSE} = \sum_{i=0}^N \vert c_i-\tilde{c}(x_i)\vert^2\).% This prediction head is trained after the encoder model has been pretrained. At this stage, we discard the projector network and use the representation \(R(x)\) as input to the prediction model. 
%In read alignment, the goal is to get the exact alignment matches, and as previously mentioned, we can use local alignment techniques to go from an approximate location to the best match in a small window. 

%As a baseline method, we consider a straightforward regression approach where the prediction network is a 3-layer MLP trained using the MSE loss to directly predict the read coordinates. % , \(L_{MSE} = \vert c(x)- MLP(R(x)) \vert^2  \). 
% the goal is achieved if we just get an approximate location within the genome.
The other alternatives use a classification approach. A natural option would be to split the reference genome into \(B\) sized bins, and train a model to predict the probability that a read falls into each bin, as in \cite{read_aln}. This approach can be effective for small genomes or when large bins are acceptable, but leads to either large bins (reducing precision) or a huge number of bins (increasing complexity) for large genomes. We therefore explore alternate strategies that should better scale to larger genomes.
Inspired by works like \citet{chen2023analogbitsgeneratingdiscrete}, we convert coordinates \(c\) into the one-hot encoding of their bitwise representation \(c_b =[t_1,t_2, \dots, t_{N_b}]\), where \(t_n \in \{0\dots b-1\}\) and \(N_b = \lfloor \log_b(L)\rfloor+1\). For example, the coordinate \(c = 20000\) would in base 2 be expressed as 
\[
20000_2 =
\left[
\begin{array}{*{15}{c}}
0 & 1 & 1 & 0 & 0 & 0 & 1 & 1 & 1 & 0 & 1 & 1 & 1 & 1 & 1 \\
1 & 0 & 0 & 1 & 1 & 1 & 0 & 0 & 0 & 1 & 0 & 0 & 0 & 0 & 0
\end{array}
\right]
.\]
We then train a classifier to predict each bit in this sequence by outputting softmaxed probabilities for each element, using the categorical cross-entropy loss.
%Instead of binning the data with a fixed width, we take the bitwise representation in base \(b\) of the coordinate \(c_b = [t_0, t_1,\dots ,t_{N_b}]\) and train a model to classify each bit. %, in the binary case \(P(t_0,t_1,\dots b_{N_b}) = [P(t_0 = 0),P(t_1 = 0) \dots P(t_{N_b} = 0);P(t_0 = 1),P(t_1 = 1) \dots P(t_{N_b} = 1)]^\top\). 
This reformulates the task from a single classification problem with \(L/B\) classes into \( N_b \) classification problems with \(b\) classes, yielding a logarithmic rather than linear scaling of the output dimension, decreasing the problem size for large genomes. %  This model uses the same MLP as the regression model, but outputs the softmaxed probabilities of each bit, and has size \([b,\lfloor \log_b(L)\rfloor + 1]\). %, and the model is trained to minimize the categorical cross entropy. % We have chosen to use base 3, since it gives the  
 % A genome of size \(10^6 \) bps and a bin size of 1000 would mean a classification problem with 1000 classes. In base 2, our approach will turn this into 20 binary classification problems, reducing the dimensionality of the output by a factor of 25. For the human genome (3 billion base pairs), the output dimension is reduced by a factor of 15625.  %Going further, using base 3 results in 13 ternary classification problems.

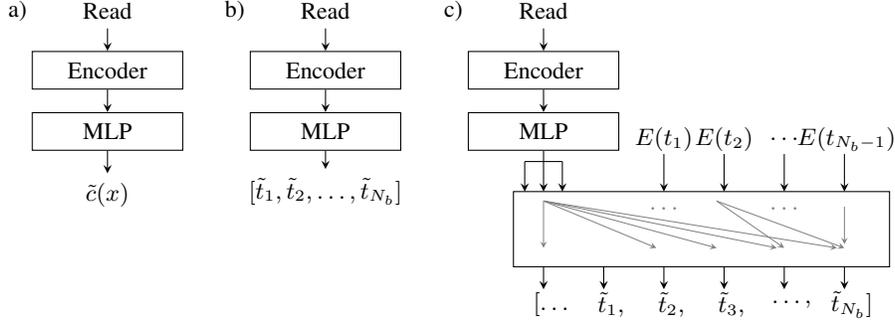
\begin{figure}[ht]
\centering
\begin{tikzpicture}[
  node distance=1cm and 1cm,
  every node/.style={font=\small},
  box/.style={draw, minimum width=2cm, minimum height=0.5cm, align=center},
  arrow/.style={->}
]

% Regression illustr.
\node[] (read1) {Read};
\node[left=0.5 of read1]  {a)};
\node[box, below=0.3cm of read1] (conv1) {Encoder};
\node[box, below=0.3cm of conv1] (mlp1) {MLP};
\node[below=0.3cm of mlp1] (coord) {\(\tilde{c}(x) \)};

\draw[arrow] (read1) -- (conv1);
\draw[arrow] (conv1) -- (mlp1);
\draw[arrow] (mlp1) -- (coord);

% Bitwise classification illustr.
\node[, right=2cm of read1] (read2) {Read};
\node[box, below=0.3 of read2] (conv2) {Encoder};
\node[box, below=0.3 of conv2] (mlp2) {MLP};

% Table grid
\node (tableOrigin) at ($(mlp2) + (-0.,-0.8)$) {\([\tilde{t}_1,\tilde{t}_2,\dots, \tilde{t}_{N_b}]\)};

\iffalse
\coordinate (tableOrigin) at ($(mlp2) + (-2.5,-0.8)$);
\foreach \i in {0,...,5} {
  \draw ($(tableOrigin)+(1*\i,0)$) -- ++(0,-1);
}
\draw ($(tableOrigin)+(0,0)$) -- ++(5,0);
\draw ($(tableOrigin)+(0,-0.5)$) -- ++(5,0);
\draw ($(tableOrigin)+(0,-1)$) -- ++(5,0);

% Table labels

\node at ($(tableOrigin)+(0.5,-0.25)$) {\(p_{0}^0\)};
\node at ($(tableOrigin)+(1.5,-0.25)$) {\(p_{1}^0 \)};
\node at ($(tableOrigin)+(2.5,-0.25)$) {\dots};
\node at ($(tableOrigin)+(3.5,-0.25)$) {\(p_{N-1}^0\)};
\node at ($(tableOrigin)+(4.5,-0.25)$) {\(p_{N}^0\)};

\node at ($(tableOrigin)+(0.5,-0.75)$) {\(p_{0}^1\)};
\node at ($(tableOrigin)+(1.5,-0.75)$) {\(p_{1}^ 1 \)};
\node at ($(tableOrigin)+(2.5,-0.75)$) {\dots};
\node at ($(tableOrigin)+(3.5,-0.75)$) {\(p_{N-1}^0\)};
\node at ($(tableOrigin)+(4.5,-0.75)$) {\(p_{N}^1\)};
\fi

\draw[arrow] (read2) -- (conv2);
\draw[arrow] (conv2) -- (mlp2);
\draw[arrow] (mlp2) -- ++(0.0,-0.5) ; %($(tableOrigin)+(0.5,0.1)$);
\node[left= 0.5 cm of read2]  {b)};

% Bitwise GPT predictor illustr.
\node[, right=2cm of read2] (read3) {Read};
\node[left=.5cm of read3]  {c)};

\node[box, below=0.3 of read3] (conv3) {Encoder};
\node[box, below=0.3 of conv3] (mlp3) {MLP};

% Shifted rectangle (mimicking right-shifted table)
\coordinate (barOrigin) at ($(mlp3) + (-0.4,-1.8)$);
\draw (barOrigin) rectangle ++(5,1);

% Labels Above bar

\node at ($(barOrigin)+(2.0,1.7)$) {\(E(t_1)\)};
\node at ($(barOrigin)+(2.0,1.7-0.925)$) {\( \color{gray}\dots \)};
\node at ($(barOrigin)+(2.8,1.7)$) {\(E(t_2)\)};
\node at ($(barOrigin)+(3.6,1.7)$) {\(\dots\)};
\node at ($(barOrigin)+(4.4,1.7)$) {\( \,  E(t_{N_b-1})\)};

% Labels below bar
\node at ($(barOrigin)+(0.5,-.5)$) {\( [\dots\)};
\node at ($(barOrigin)+(1.3,-.5)$) {\(\tilde{t}_1,\)};
\node at ($(barOrigin)+(2.1,-.5)$) {\(\tilde{t}_2,\)};
\node at ($(barOrigin)+(2.9,-.5)$) {\(\tilde{t}_3,\)};
\node at ($(barOrigin)+(3.7,-.5)$) {\(\dots,\)};
\node at ($(barOrigin)+(3.6,1.7-0.925)$) {\( \color{gray}\dots \)};
\node at ($(barOrigin)+(4.5,-.5)$) {\(\tilde{t}_{N_b}]\)};

 %%Arrows from MLP to bar
\foreach \x in {  3.4, 4.2,5.0,5.8} {
  \draw[arrow] (mlp3) ++(\x-1.8,-0.3) -- ++(0,-0.5);
}

\foreach \x in { 2.6, 3.4, 4.2,5.0,5.8} {
  \draw[arrow] (mlp3) ++(\x-1.8,-1.8) -- ++(0,-0.3);
}
\draw[arrow] (mlp3) + (0,-1.8) -- ($(barOrigin)+(0.4,-0.3)$);

% Arrows within GPT
\foreach \x in { 1.9, 3.4, 4.2,5.0,5.8} {
  \draw[arrow, color=gray, line width=0.1pt] (mlp3) ++(0,-1.125+0.2) -- ++(\x- 1.9,-0.725+0.1);
}

%\foreach \x in {   4.2,5.0,5.8} {
%  \draw[arrow,line width=0.1pt, color=gray] (mlp3) ++(1.5,-1.125) -- ++(\x- 3.3,-0.625);
%}

\foreach \x in {  5.0,5.8} {
  \draw[arrow,line width=0.1pt, color=gray] (mlp3) ++(2.3,-1.125+0.2) -- ++(\x-  4.1,-0.725  +0.1);
}

\draw[arrow,line width=0.1pt, color=gray] (mlp3) ++ (4.0,-1) -- ($(barOrigin)+(4+0.4,0.3)$);
%\node at ($(barOrigin)+(4.4,1.7)$) {\( \,  E(b_{N-1})\)};
%\draw[arrow] (mlp3) + (0,-2) -- ($(barOrigin)+(0.4,-0.4)$);

%\foreach \x in { mlp3, 3.4, 4.2, 5.0, 5.8} {
%  \draw[arrow] (mlp3) ++(\x-1.8,-0.1 -1.4) -- ++(0,-0.4);
  %([shift={(\x,-0.1)}]) -- ++(0,-0.3);
  %\draw[->, thick] ([shift={(\x-1.8,-0.1 -1.4)}]mlp3.south) -- ++(0,-0.4); 
%  } 

\draw[arrow] (read3) -- (conv3);
\draw[arrow] (conv3) -- (mlp3);
%\draw[arrow] (mlp3) -- ($(barOrigin)+(0.4,1.2)$);

% Unified arrow splitting into three: horizontal then down
\coordinate (arrowStart) at ($(mlp3)+(0,-0.4)$);
\coordinate (splitLeft)  at ($(arrowStart)+(-0.25,0)$);
\coordinate (splitMid)   at ($(arrowStart)+(0,0)$);
\coordinate (splitRight) at ($(arrowStart)+(0.25,0)$);

% Vertical from MLP to split line
%\draw[arrow] (mlp3) -- (arrowStart);

% Horizontal split line
\draw[] (splitLeft) -- (splitRight);

% Downward arrows from split points
\draw[arrow] (splitLeft) -- ++(0,-0.4);
\draw[arrow] (mlp3) -- ++(0,-0.8);
\draw[arrow] (splitRight) -- ++(0,-0.4);

\end{tikzpicture}

\caption{Different prediction heads for the coordinate prediction task. a): An MLP using regression to predict \(k\)-mer position \(\tilde{c}(x)\). b): MLP predicting the bitwise probabilities using the representation \(h\). c): A small GPT predicting probabilities using \(h\) and the previous bits.  }
\label{fig:pred_heads}

\end{figure}

In the above model, all bits are predicted independently, and it does not utilize the fact that the value of a more significant bit would influence the interpretation, and hence the predictions, of subsequent bits. To incorporate this information, the last approach we introduce is a bit-predicting GPT model. 
%However, there is structure in the bitwise representation that is not utilized in this approach. The value of each bit has information that the subsequent bit predictions would benefit from, but in this plain classifier approach, all bits are predicted independently.
%As a third prediction head we use a  GPT-like architecture. 
We employ a lightweight GPT that follows the standard transformer architecture  \cite{Radford2018ImprovingLU}, consisting of causally masked multi-head self-attention, layer normalization, and a feed-forward network. The model uses just one transformer block, 2 heads, a feed-forward dimension of 256, and a token dimension of 64. We set the output dimension of the MLP to 512, and split it into 8 64-dimensional tokens, which become the starting sequence during inference. We use a trainable embedding for the bit-tokens and add a trainable positional encoding. The model has full attention across the MLP tokens, and each bit prediction uses information from the MLP tokens, \textit{and only} the previously predicted bits, increasing the information used in each prediction. The prediction models are illustrated in  Figure \ref{fig:pred_heads}.

In read alignment, we want to get the exact placement of the reads within the reference, and predicting the exact position proves challenging for any of these methods. 
If we can predict an approximate position, local alignment can be performed cheaply by sliding the read across a small window near the predicted location and finding the best match. This local alignment can be efficiently parallelized on a GPU, while attempting to do this over the full genome is computationally infeasible. Classical algorithms for alignment, such as BWA, also tend to first identify a seed position and then perform an alignment/extension process around that seed. For short reads, the seed identification process dominates the computational load.

%, and we instead aim to predict an approximate position, after which local alignment can be performed. This refinement step is done by sliding the read across a small window near the predicted location and finding the best match. This short-window local alignment can be efficiently parallelized on a GPU, while attempting to do this over the full genome is computationally infeasible. %This means we are happy with only a coarse grained prediction.

\paragraph{Training and Evaluation data.}
To illustrate the potential of our method, we apply it to learn the structure of the \textit{Escherichia coli (E.\ coli)} genome, accession number NC\_000913 \cite{NC_000913_3}. It has 4.64 million nucleotides and is a common model species.
In the numerical experiments, we use the reference genome as the starting sequence.  We draw $k$-mers with $k=30$, and use a maximum positive offset of \(d = 50\). % To simulate From this, we draw \(N\) starting positions from \(\mathcal{U}(0,L)\). For each of them, we also draw an explicit positive sample at most 30 bp away to  ensure that there is always a positive sample for each anchor. 
%Then we apply the aDNA noise model which consists of a background noise level of \(1\%\), and by increasing the \(C\to T\)  and \( G \to A\) substitutions for the first and last ten bases to \(10\%\), respectively.  Lastly, we apply the complement augmentation by flipping both axes with probability 0.5 to ensure that the model is robust with respect to the complement. 
When evaluating model performance on unseen data, we create 30bp reads using the Gargammel ancient DNA simulator \citep{görgammel}, which simulates Illumina sequencing errors using the ART read simulator \cite{art_readsim} and applies aDNA noise using the Briggs model \citep{briggs}.  

%We summarize the pipeline in Algorithm \ref{alg:pipeline}

\section{Numerical Experiments}
\begin{figure}[htb]
    \centering
    \includegraphics[width=0.99\linewidth]{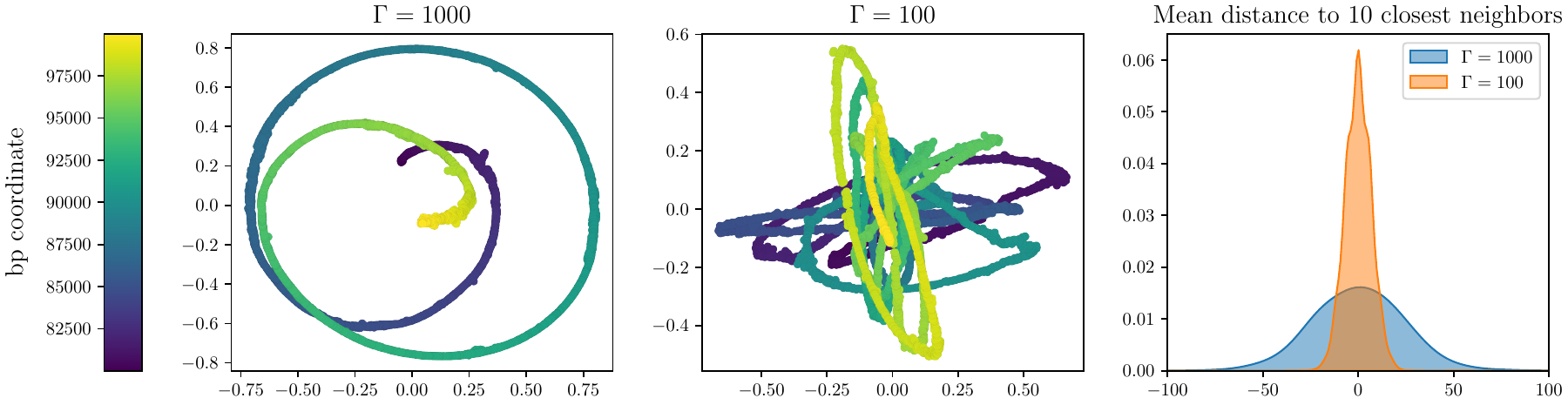}
    \caption{2D PCA of the resulting embedding over a 20kbp window, with \(\Gamma = 1\,000\) (left) and \(\Gamma=100\) (middle), and the distribution of the mean distance to the 10 closest neighbors for each sample (right).}
    \label{fig:emb}
\end{figure}

In this section, we evaluate how well the embeddings reflect genomic structure, present the short-read mapping performance, and demonstrate other potential uses of a pre-trained model. When visualizing \(k\)-mer embeddings, we take a small excerpt of the genome since the full embeddings are very long trajectories in \(\mathbb{R}^{256}\) which are hard to visualize in 2D plots. 
\paragraph{Genome Embeddings and Thresholding Parameter.}
Here we study the effect that the magnitude of \(\Gamma\) has on the embedding.
We train the CReadNet-B on the first \(10\%\) of the \textit{E.\ coli} genome, and Figure \ref{fig:emb} shows a scatterplot of a 20kbp snippet of the embedding, where each marker is an embedded \(k\)-mer, colored by its true bp coordinate.  We have taken the normalized 256-dimensional embedding and applied PCA to visualize it in two dimensions.
The subpanels of Figure \ref{fig:emb} show, from left to right: the embedding using \(\Gamma = 1\,000\), the embedding using \(\Gamma = 100\), and a density plot of the mean distance to each embedded \(k\)-mer's 10 closest neighbors.
%The left image shows the embedding using \(\Gamma = 1\,000\), the middle one using \(\Gamma=100\), and the right image is a density plot of the mean distance to each embedded \(k\)-mer's 10 closest neighbors. 
Both choices of \(\Gamma\) produce embeddings that preserve the genomic structure, with a trajectory reflecting the ordering of the coordinates of the \(k\)-mers, and both have
an average bp distance to the 10 closest \(k\)-mers less than 100.  \(\Gamma\) determines how tightly the reads will be mapped, with a smaller \(\Gamma\) better capturing the local structure, while a larger value gives a less complex manifold, which may simplify downstream tasks.

\begin{figure}[htb]
    \centering
    \includegraphics[width=1\linewidth]{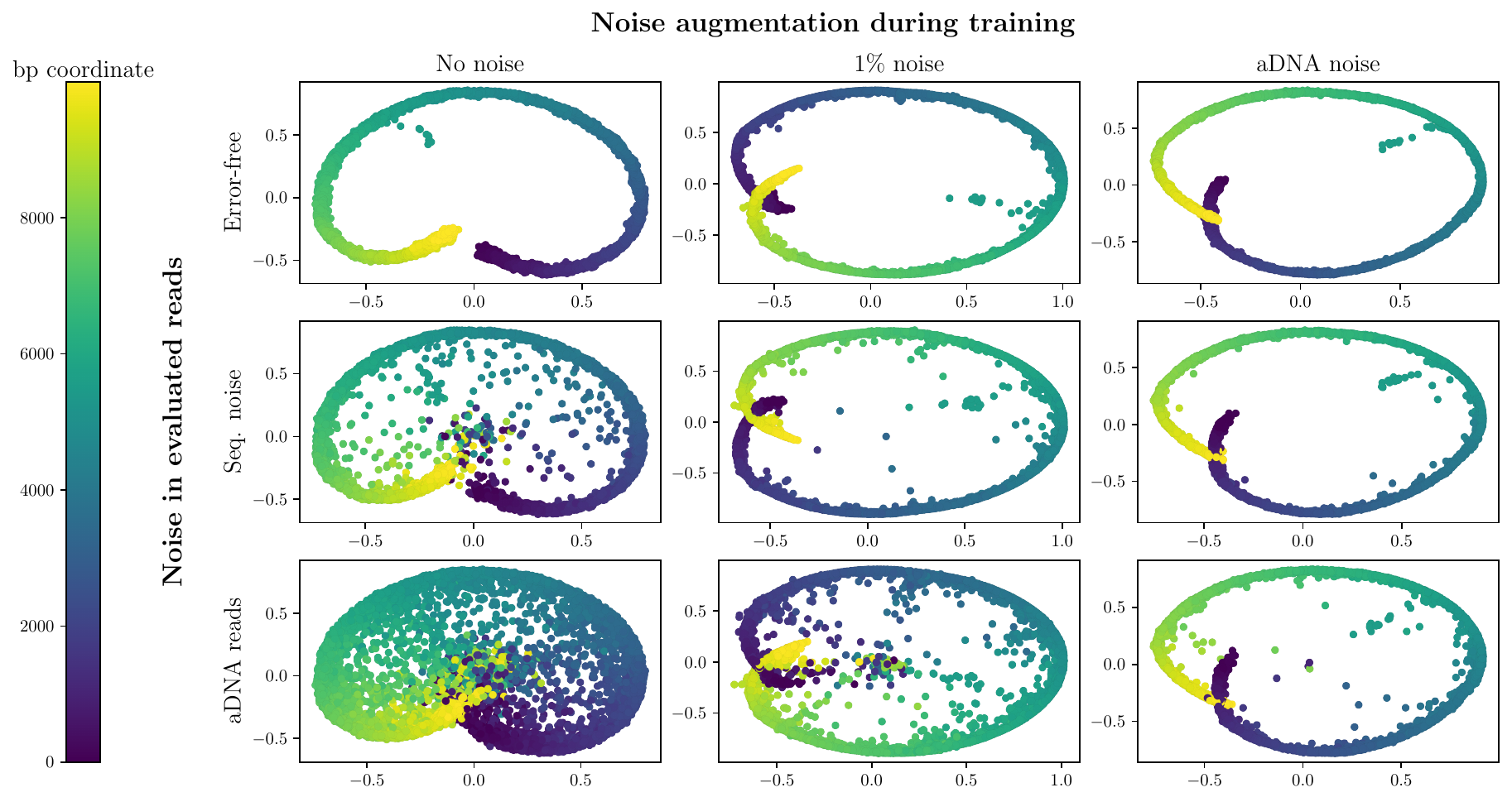}
    \caption{2D PCA plots of the first 10k base pairs of embeddings, showing the difference in the resulting embedding when training with different augmentation strategies (columns), for varying levels of data corruption (rows).}
    \label{fig:noise}
\end{figure}

\paragraph{Better Error Model Improves Performance.}
Here we show the influence of the augmentation scheme, specifically, which type of noise is applied. Figure \ref{fig:noise} shows a 2D PCA plot of the first 10k base pairs using CReadNet-T, trained on 10\% of the \textit{E.\ coli} genome. Across columns, the plots show the augmentation scenarios of no added noise, a flat 1\% substitution rate, and the aDNA error noise model, respectively. Across rows, the evaluation data is instead varied, with error-free reads, simulated reads using only sequencing noise, and simulated aDNA reads using Gargammel, respectively. The clean upper triangle shows that a model trained on high noise generalizes well to less noisy data, but not vice versa.

\paragraph{Read Alignment of aDNA Reads.}
Given trained embeddings, we train the three read mapping prediction heads for both the first \(10\%\) and the full \textit{E.\ coli} genome, and evaluate the mapping performance on simulated aDNA reads. We denote the regression model \textit{MSE}, the bit prediction model \textit{CCE}, and the GPT model \textit{GPT}, all trained for 10k iterations, and use base 3 in the bitwise representation. The mapping results are shown in Table \ref{tab:mapping}, where we report the percentage of correctly mapped reads after the local alignment step, and the throughput measured in millions of reads processed per second. We compare with BWA-aln, with different values of the maximum edit distance \(n\). The default value, which is also recommended in the literature for aDNA, corresponds to \(n=3\)  for 30 bp reads \cite{adna_bwa_aln}. BWA-mem is usually considered the standard method for short-read alignment, but often underperforms for aDNA \citep{adna_bwa_aln}. Bowtie2 \cite{bowtie} is another common read aligner, also for aDNA, but we include both as a reference for standard aligners that fail when aligning aDNA reads.  For BWA-mem and Bowtie2 we have used the best performing (BP) parameters for aDNA \cite{adna_bwa_aln,Oliva2021}. %For reference, BWA-mem only correctly maps 54\% of the reads in the 10\% case with default settings.
Reported scores are the medians across 3 different seeds. The scores for each run are shown in the Appendix, as well as plots illustrating the performance without local alignment. The reported throughput for our models is measured by averaging 10 steady-state inference steps. % skipping the first kernel-building step.
The timings for the local alignment step are omitted. A poorly optimized CuPy \cite{cupy} implementation we made has a throughput of 0.68 million reads per second on a single A100 when using a 5000bp window, and better optimized approaches should yield negligible overhead.

\begin{table}[htb]%{\linewidth}
\caption{Ancient DNA read mapping score reported as the percentage of correctly mapped reads, along with the throughput of the methods, reported in \(10^6\) reads/s (MR/s). The mode denotes the number of allowed mismatches for BWA-aln, or the prediction head used. Best practice parameters are used for Bowtie2 and BWA-mem.}

 \label{tab:mapping}
 %\begin{tabular}{llllllll}
   \centering

 \begin{tabular}{lccc@{}c@{}ccccccccccc}
    \toprule
    \multicolumn{1}{c}{ } & \multicolumn{3}{c}{10\% } &   \multicolumn{1}{c}{ }&  \multicolumn{9}{c}{100\%} \\
    \cmidrule(r){2-4}
    \cmidrule(r){6-14}
    %\cmidrule(r){7-8}
    %\cmidrule(r){9-11}
    %Method     & Setting     & acc \(\%\) & M reads/s & Method     & Setting     & acc \(\%\) & M reads/s \\
        Method     & Mode  & Acc. &  MR/s & \hspace{1.em} & Mode & Acc.  & MR/s & Mode  & Acc. &  MR/s & Mode  & Acc. &  MR/s  \\
        %\([\%]\)
    \midrule

    %BWA aln& \(n=1\) &84.374  & 624\,285&BWA aln& \(n= 1\) &83.983 &  711\,118\\
    %BWA aln& \(n=2\) &94.937  &  521\,548& BWA aln& \(n= 2\) &94.180 &  458\,672\\
    %BWA aln& \(n=3\) &98.018  &375\,325&BWA aln& \(n= 3\) &97.078 & 230\,326\\
    %BWA aln& \(n=4\) &98.798 &  241\,510& BWA aln& \(n= 4\) &97.832 &  117\,191\\
    %BWA aln& \(n=5\) &98.996 & 158\,056&BWA aln& \(n= 5\) &98.011 &  65\,379\\
    %BWA aln& \(n=6\) &99.056 & 114\,331&BWA aln& \(n= 6\) &98.056 &  42\,299 \\
    %Ours T & \(\Gamma=1000 \) &   389\,896 &Ours T  & \(\Gamma=1000 \) & \(93.69^*\)& \\
    %Ours S &\(\Gamma=1000 \) &    202\,712&Ours S  & \(\Gamma=1000 \) & \(97.24^*\) & \\
    %Ours B  &\(\Gamma=1000 \) &   124\,673 &Ours B  & \(\Gamma=1000 \) & 98.03 & \\
    
    %BWA aln& \(n=1\) &84.374  & 624\,285 &83.983 &  711\,118\\
    %BWA aln& \(n=2\) &94.937  &  521\,548 &94.180 &  458\,672\\
    %BWA aln& \(n=3\) &98.018  &375\,325 &97.078 & 230\,326\\
    %BWA aln& \(n=4\) &98.798 &  241\,510 &97.832 &  117\,191\\
    %BWA aln& \(n=5\) &98.996 & 158\,056 &98.011 &  65\,379\\
    %BWA aln& \(n=6\) &99.056 & 114\,331 &98.056 &  42\,299 \\
    %Ours T & \(\Gamma=1000 \) &     &389\,896 & \(93.69^*\) \\
    %Ours S &\(\Gamma=1000 \) &   &  202\,712  & \(97.24^*\) \\
    %Ours B  &\(\Gamma=1000 \) &   & 124\,673 &98.03 \\
    Bowtie2  &BP &  9.78 & 0.023& & BP &  13.14  &0.023&  -- &-- & --& --& --& -- \\
    BWA-mem & BP &54.76 &0.204 & & BP &  54.89  & 0.181 &  -- &-- & --& --& --& -- \\
    %BWA-aln& \(n=1\) &84.37  & 0.624 & &\( n=1\)&83.98 &  0.711 &  -- &-- & --& --& --& -- \\
    %BWA& \(n=2\) &94.937  &  0.521 &94.18 &  0.458& -- &-- & --& --& --& -- \\
    BWA-aln& \(n=3\) &98.02  &0.375& & \(n=3\)&97.08 & 0.230 & -- &-- & --& --& --& -- \\
    %BWA& \(n=4\) &98.798 &  0.241 &97.83 &  0.117& -- &-- & --& --& --& -- \\
    %BWA& \(n=5\) &98.996 & 0.158 &98.01 &  0.065& -- &-- & --& --& --& -- \\
    BWA-aln& \(n=6\) &99.06 & 0.115&  & \(n=6\)& 98.06 &  0.042 & -- &-- & --& --& --& --  \\
    CReadNet-T & \textit{CCE}&  98.71   & 0.424 & & \textit{CCE}&  94.19 & 0.422 & \textit{GPT} &95.26 &  0.117 &  \textit{MSE}  &14.40 &  0.426\\
    CReadNet-S & \textit{CCE} &  98.78 & 0.232 &   & \textit{CCE} & 95.72&  0.229  & \textit{GPT} &96.59&  0.095&  \textit{MSE} & 21.50 &  0.231\\
    CReadNet-B  & \textit{CCE} &   98.78 & 0.139 &  & \textit{CCE}& 97.47 &  0.137& \textit{GPT} & 97.66&  0.074 &  \textit{MSE}  & 65.64 &  0.138\\
    %Ours B  & GPT &  99.01 & XXX  & 97.411  &  0.0739 \\ %\(\pm 0.01126\) 
    \bottomrule
  \end{tabular}
\end{table}

\begin{figure}[htb]
    \centering
    \includegraphics[width=0.99\linewidth]{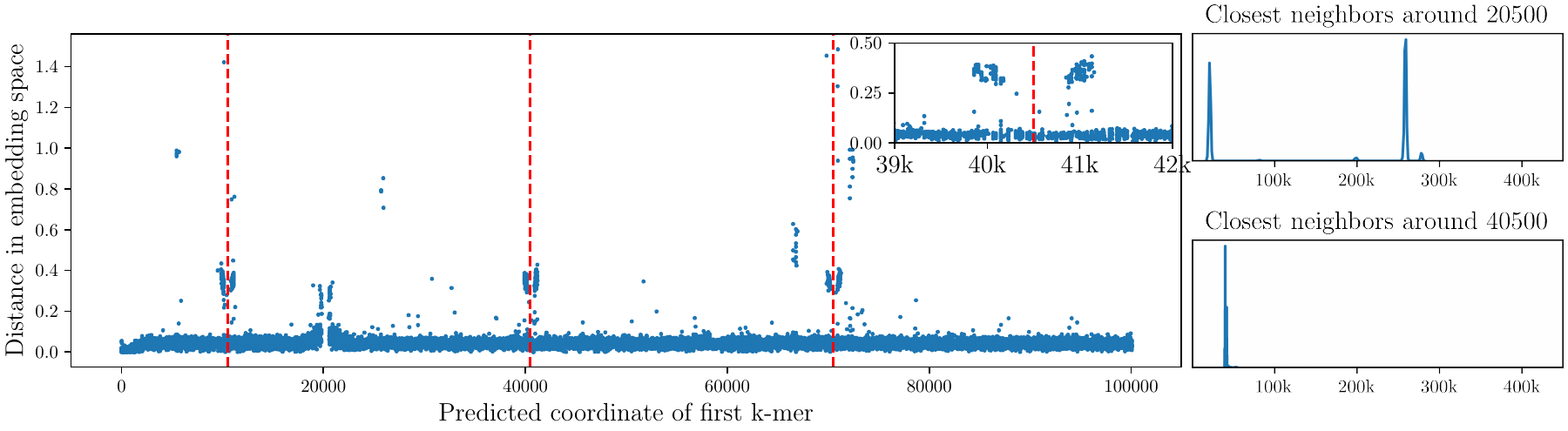}
    \caption{The distance in embedding space between two 30bp k-mers from the same 150bp read, plotted against the predicted coordinate, indicates structural variation. The distances highlight both artificially added inversions (red dashed lines), as well as duplications found in the \textit{E.\ coli} genome.}
    \label{fig:inversion_detection}
\end{figure}

\paragraph{Inversion Detection and Structural Variation.}
%We have now shown that the embeddings can be used to train prediction models to map noisy reads. 
%As mentioned, the pre-trained encoder model gives a general representation of the genome and can be used other other task than read alignment. %, and can be used for  is not tailored specifically for this task, and can be used for other tasks as well. 
In addition to substitutions, insertions, and deletions, \textit{inversions} can also occur, where a section of the genome has been replaced by its reverse complement, such as \( \text{TG\textbf{C...GT}GG} \to  \text{TG{\color{red}AC...G}GG}\). Such inversions may have a detrimental effect if certain genes are affected. % and the detection can be important.
%We can use our trained encoder models to detect inversions from longer reads.
Since the encoder maps short subsequences close to each other, the distance in embedding space between two \(k\)-mers from the same longer read should be small. If, however, a pair of \(k\)-mers come from different sides of the border of an inversion, we would expect an abnormally large embedding distance. To illustrate how our embedding can be used for inversion detection, we train CReadNet-T on 10\% of the \textit{E.\ coli} reference genome. Then, we create a new reference with inversions 1kbp in length at coordinates 10k, 40k, and 70k, and generate 150bp reads using the ART read simulator \cite{art_readsim}.
%Since sequencing methods are agnostic to which strand of the DNA the reads come from, you would only be able to detect inversions from reads that cover one end of the inversion, since reads from inside an inversion look just like the complement of the read. 
%
%To illustrate this, we insert three inversions into the first 10\% of the e-coli genome, generate 150bp long reads. 
From these reads, we draw two \(k\)-mers as the first and last 30 bps, and the pairwise embedding distance is shown in  Figure \ref{fig:inversion_detection}. The three inversions stand out, but also another region at around 20kbp. 
The right-hand subpanels show the distribution of predicted coordinates of the 250 closest embedding neighbors for \(k\)-mers predicted to be inside the proposed inversion region. 
%The right plots show the distribution  of samples predicted close to the detected structural variation points 40500 and 20500. 
%The samples are counted as follows: Take all samples predicted to be within 250 bps of the inversion point, look at their 100 closest neighbors in embedding space, and plot the distribution of their predicted coordinates. 
From this, we can infer whether we have a duplication within the genome or a structural variation in the form of an inversion. If we have a multimodal distribution, it is a repeating sequence found at multiple points in the genome, and if it is unimodal, we can conclude it is an inversion.

\begin{figure}[htb]
    \centering
    \includegraphics[width=0.99\linewidth]{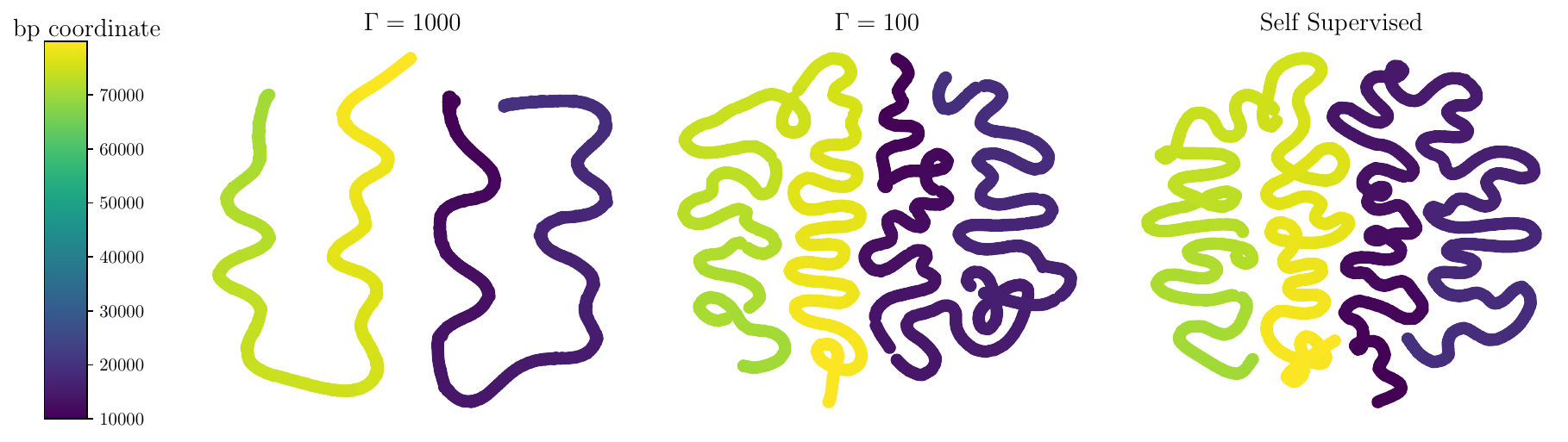}
    \caption{UMAP applied to embedding produced by CReadNet-T trained on two 10kbp excerpts from the \textit{E.\ coli} genome, using \(\Gamma= 100\), \(\Gamma= 1\,000\), and using a self-supervised approach mimicking  a case where no reference is available.}
    \label{fig:2slice}
\end{figure}

\paragraph{Disjoint Sequences and Non-Model Species.}
There are organisms for which we do not have a reference. In that case, we are unable to use the supervised contrastive framework, but are limited to training on reads only. This is simulated by training solely on pairwise augmented \(k\)-mers as positives. The data used in this experiment is two disjoint 10k subsequences of the \textit{E.\ coli} genome starting at bp 10k and 70k, simulating a case with data from two different organisms.
In Figure \ref{fig:2slice} we show 2D UMAP \cite{umap} plots of the embedding \(  \Gamma= 1\,000\), \(\Gamma= 100\), and the self-supervised loss. % which is obtained by simply removing the positive samples defined based on \(\Gamma\), and the distance scaling term. 
In all three cases, we can see that the embedding reflects that the two sequences are disjoint. Clustering \(k\)-mers in two groups is trivial. For longer sequences, this methodology can be used for metagenomic species identification. This has been run on a small toy example for ease of visualization as a proof of concept.

\section{Conclusions}

In this paper, we have introduced a framework for learning the genomic structure from short sequences of DNA called \(k\)-mers. Previous literature on using deep learning methods for analyzing DNA data has often been developed for a single  application in mind, whereas we in this work produce a general representation that can be used for several applications. 

We introduce the bitwise prediction of the read coordinates. This approach offers several benefits. A bitwise representation can be thought of as a hierarchical binning approach. In base two, the most significant bit decides in what half of the reference the \(k\)-mer should be placed, corresponding to two bins of size \(L/2\). Given the first bit, the next bit determines the next halving of the interval, a classification of two bins of size \(L/4\), and so on. The bit representation reduces the number of outputs while still allowing for fine-grained predictions.
Another advantage over a direct regression output is that the model outputs softmaxed probabilities, giving quantified confidences for each predicted bit. These could be used to make the local alignment step even cheaper by adapting the window size in the local alignment step. Predicting each bit in isolation works surprisingly well, outperforming the regression approach with essentially the same model complexity. We show that the GPT can utilize this and attain even higher accuracy -- albeit at a larger inference cost. This approach could be extended to other large-domain prediction tasks to reduce complexity. %An example could be 

Table \ref{tab:mapping} shows that the prediction models perform on par with the default values for the gold-standard BWA-aln in aDNA mapping. Setting the maximum edit distance to 6 increases the accuracy, but it is not a standard choice in the literature. BWA-aln reports that 3.5\% of all reads from the full \textit{E.\ coli} genome are ambiguous, with multiple hits across the genome. With this in mind, accuracies above 96.5\% will only improve by choosing the correct position out of a set of equally good alignments.
BWA-aln scales logarithmically with the size of the reference, whereas for a fixed model complexity, inference using our model stays essentially constant. % , needing only to predict a couple more bits. 
%we still think this shows that our method hold promise when it comes to inference speed, with BWA-aln scales with the size of the reference. 
Also, there is performance to be gained by using model quantization and weight pruning, which we have not explored so far. %. Comparing full end-to-end read mapping of a new reference we can not compete due to the expensive contrastive pre-training. This only has to be done once, after which we have a representation that is not limited in use to mapping.

Thus far, we have considered the reference genome as the ground truth for the genomic structure of a species. That is an idealization that has been common in the field of genomics. However, the established reference genomes can lack genomic regions, including full genes, carried by some individuals within a species. There can also be major structural variations, where the same genes can appear in different distinct orderings. This is an issue for bacterial genomes \cite{miss_bact}, but it is also of significant importance in the human reference GRCh38, which misses up to millions of base-pairs \cite{Shi2016,Ameur2018-na,Wu2024-hy}. Since our method can operate without a reference, it can recognize the full genomic complexity, as long as it is present in the training data. In the specific case of aDNA mapping, so-called reference bias \cite{Gunther2019-pc} is also an important issue. When considering single letter genetic variants (single nucleotide polymorphisms, SNPs), reads that contain the reference allele are successfully mapped more frequently. This may have a detrimental effect on downstream analyses. In our framework, this could be amended by applying SNP-specific augmentation schemes that would ensure that alternate alleles would map just as well as their reference counterparts. For both of these applications, there is also an equity aspect, since many reference resources were originally defined based on privileged populations found in developed nations. Likewise, for studying issues such as antimicrobial resistance, such genes can appear in regions that deviate from the reference sequence.

\paragraph{Future work.}
Figure \ref{fig:2slice} demonstrates two possible future applications. Our model produces embeddings of \(k\)-mers that distinguish samples coming from two disjoint sequences, simulating a case where we have reads from two different organisms. Here, reads from different organisms, or different chromosomes or plasmids within an organism, would form distinct trajectories in the embedding, enabling metagenomic species identification based on \(k\)-mers. 
It also shows that self-supervised training could be used for \emph{de novo} assembly, building up the full genome sequence from reads into a trajectory in the embedding space, from which a skeleton graph can be extracted. 
%In this case, it was a toy example with to isolated parts of the \textit{E.\ coli} genome, but . It also shows that, while \(\Gamma\) can be used to make embeddings tighter, the model can be trained fully self-supervised as well, which means that we can train on reads from organisms which has no reference and use the method for assembly.

\paragraph{Limitations.}
Currently, we have only trained the model on a bacterial genome. Scaling the model up to e.g. human genomes (from $5\cdot10^6$ to $3\cdot10^9$ bps) will pose additional challenges. Training the embedding step is highly expensive, compared to building the index for conventional mapping methods, while our approach might be competitive compared to performing a full \emph{de novo} genome assembly. The training cost might be decreased with transfer learning from a pre-trained model to a related genome.

\bibliographystyle{unsrtnat}

\bibliography{references}

%%%%%%%%%%%%%%%%%%%%%%%%%%%%%%%%%%%%%%%%%%%%%%%%%%%%%%%%%%%%

\appendix

\section{Technical Appendices and Supplementary Material}
% Technical appendices with additional results, figures, graphs and proofs may be submitted with the paper submission before the full submission deadline (see above), or as a separate PDF in the ZIP file below before the supplementary material deadline. There is no page limit for the technical appendices.

\subsection{Additional position-prediction results}

In addition to the mapping scores shown in Table \ref{tab:mapping}, this section provides additional details on  the raw accuracy of the position prediction heads before, when the local alignment step is omitted.
An effective mapping model assigns many reads close to their true locations. To quantify this, we compute the empirical cumulative distribution function (eCDF) of the absolute prediction error $\vert \tilde{c}(x_i) - c_i \vert$, representing the fraction of reads mapped within a given distance from their true positions. That is, we compute \begin{equation} eCDF(t) = \frac{1}{N} \sum_{i = 1}^N \mathds{1}_{\vert\tilde{c}(x_i) - c_i\vert < t} \end{equation}

\begin{figure}
    \centering
    \includegraphics[width=0.99\linewidth]{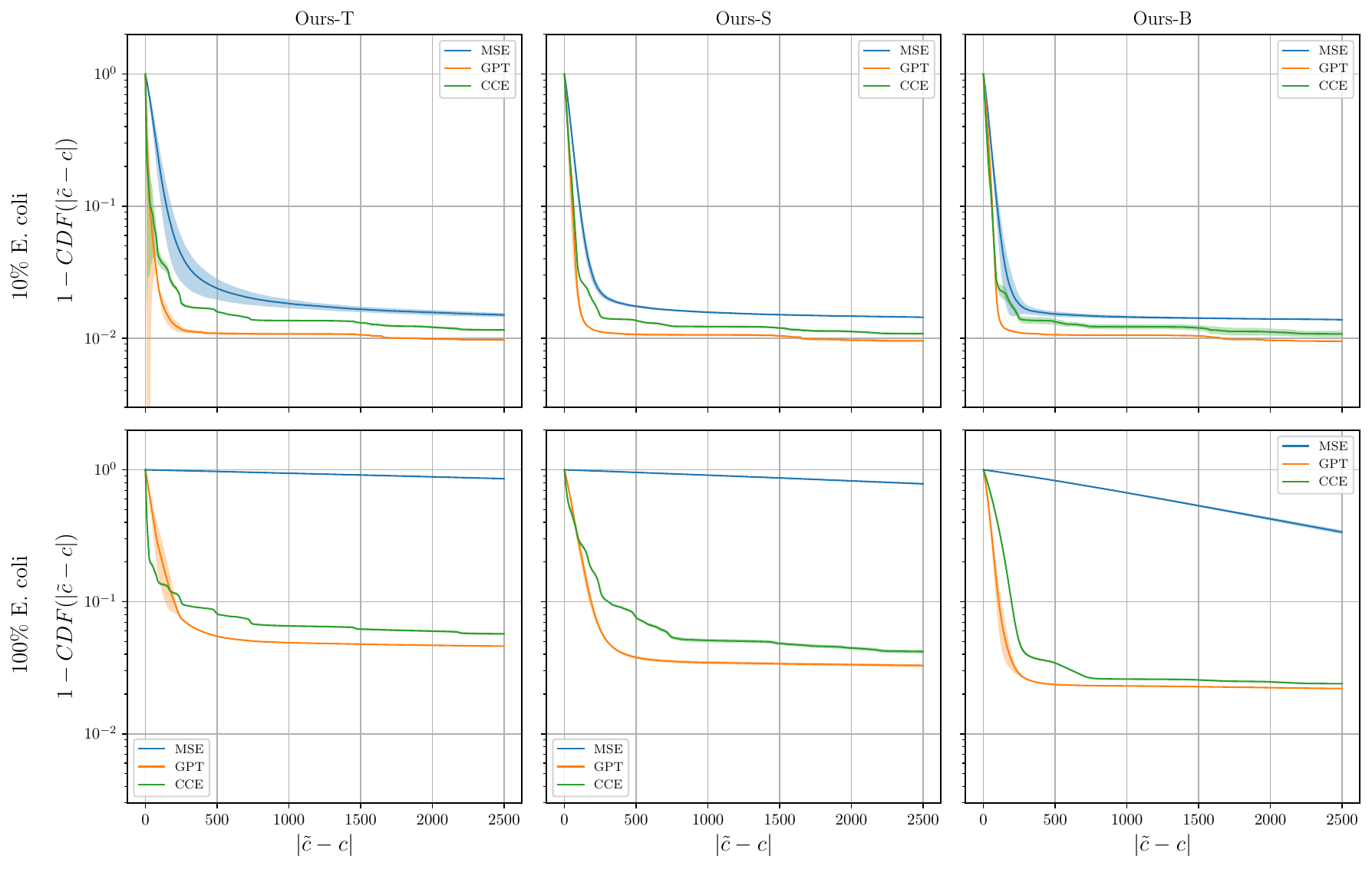}
    \caption{Performance of the different models measured by the quantity empirical CDF of the prediction error. Shown is 1-CDF, lower is better. }
    \label{fig:all_logged}
\end{figure}

\begin{figure}
    \centering
    \includegraphics[width=0.99\linewidth]{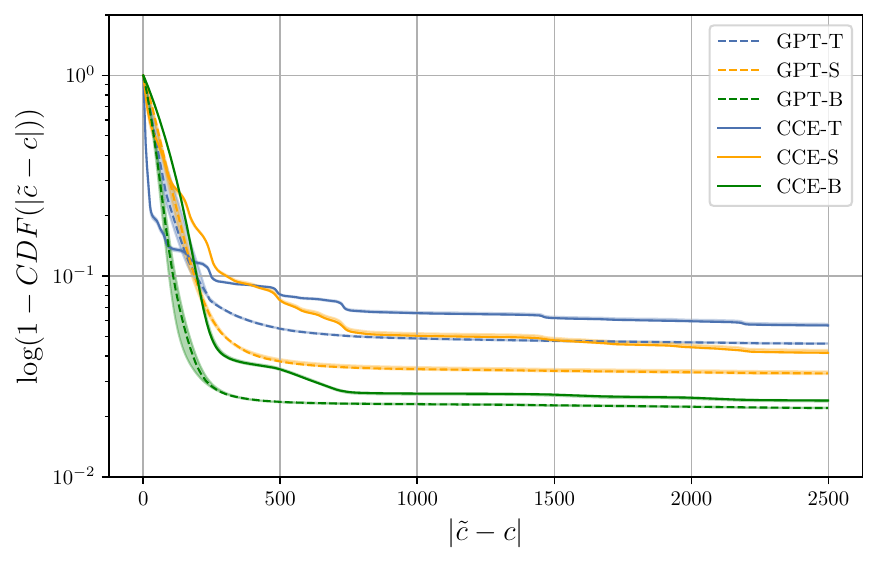}
    \caption{Closer look at the prediction performance on the full \textit{E.\ coli} with the two bit-prediction models and three different sizes of CReadNet. }
    \label{fig:GPT_vsCCE}
\end{figure}

In order to amplify differences of importance, we plot \(1- eCDF(t)\) in a logarithmic scale for the runs on 10\% and 100\% of the \textit{E.\ coli} genome, for all models, in Figure \ref{fig:all_logged}. For each model we made three replicate runs. The lines in the plots are the median, and the shaded region around each represents the minimum and maximum. In many of the plots, the shaded regions are barely visible, reflecting the high reproducibility for the bit-predicting models. We opted to report the full span of our replicates centering on the median, rather than standard deviations, due to the low sample count. This, in turn, was limited due to available computational resources. Figure \ref{fig:GPT_vsCCE} compares the two bit-prediction models with different sizes of CReadNet trained on the full genome, showing that the more complex models produce embeddings that enable the prediction heads to predict the coordinate more accurately. It also shows that in terms of precision, the GPT model outperforms the base MLP.  

The post-alignment scores for all runs are shown in Table \ref{tab:model-scores}.

\begin{table}[htb]
    \centering
    \begin{tabular}{lllll}
    \hline
    \textbf{Data size} & \textbf{Model} & \textbf{Min} & \textbf{Median} & \textbf{Max} \\
    \hline
   & Tiny-MSE & 98.36 & 98.36 & 98.41 \\
   & Tiny-CCE & 98.70 & 98.71 & 98.73 \\
   & Tiny-GPT & 98.89 & 98.90 & 98.91 \\
   & Small-MSE & 98.43 & 98.43 & 98.44 \\
10\%   & Small-CCE & 98.78 & 98.78 & 98.80 \\
   & Small-GPT & 98.90 & 98.91 & 98.93 \\
   & Big-MSE & 98.49 & 98.49 & 98.50 \\
   & Big-CCE & 98.77 & 98.78 & 98.83 \\
   & Big-GPT & 98.91 & 98.92 & 98.92 \\
    \hline
   & Tiny-MSE & 14.24 & 14.40 & 14.55 \\
   & Tiny-CCE & 94.09 & 94.19 & 94.20 \\
   & Tiny-GPT & 95.23 & 95.26 & 95.27 \\
   & Small-MSE & 21.03 & 21.50 & 22.40 \\
100\%   & Small-CCE & 95.58 & 95.72 & 95.73 \\
   & Small-GPT & 96.51 & 96.59 & 96.62 \\
   & Big-MSE & 65.59 & 65.64 & 66.91 \\
   & Big-CCE & 97.47 & 97.47 & 97.50 \\
   & Big-GPT & 97.65 & 97.66 & 97.67 \\
    \hline
    \end{tabular}
    \caption{The scores for the replicate runs made, the same as used in reporting the median in Table \ref{tab:mapping}.}
    \label{tab:model-scores}
\end{table}

%\subsection{Comparison of Prediction Heads}
%Here we compare the performance of four different ways of predicting the read position given the representation: regression training using MSE, MAE,. Here we use the same architecture for the MLP model preceding the final layer: 3 dense layers 1024 nodes,  SiLU activation, and Layernorm before all dense layers. The prediction heads are trained for 3k iterations with a batch size of 2048.
%The results are presented in Figure \ref{fig:pred_heads_results}. 
%We see that using the binary representation drastically improves the precision in the predictions, with many more reads mapped with a smaller error.

\subsection{Training setting and hyperparameters}
Table \ref{tab:setting_table} shows the hyperparameter settings used for training the models. We used the same optimizer (AdamW) for training the encoder as well as the prediction models. We tried others (LARS, RMSprop, SGD with momentum), but found AdamW to be the most reliable across scenarios. 

\begin{table}[htb]
    \centering
    \begin{tabular}{ll}
    \toprule
         \textbf{Setting} & \textbf{Value}  \\
         \midrule
         Optimizer& AdamW, \(\beta_1 = 0.9,\beta_2 = 0.999 \)\\
         Learning rate & 0.5e-3\\
         Weight decay & 1e-5 \\
         LR Schedule &Cosine decay \\ 
         Warmup & 2500 iterations \\
         Initialization & TruncatedNormal(0, 0.05) \\
         Batch size & 4096 \\
         Epochs 10\%/100\% \textit{E.\ coli} & 50\,000/15\,0000\\
         Epochs prediction head & 10\,000\\
         %Epochs 100\% \textit{E-coli} & 150000\\
         Threshold \(\Gamma\) & 1\,000 (unless otherwise specified)\\
         Positive offset \(d\) & 50\\
        Flat substitution rate & 1\%\\
        Temperature \(\tau\) & 0.1 \\ 
         
         \bottomrule
    \end{tabular}
    \caption{Hyperparameter settings used in training both the encoder model and prediction heads.}
    \label{tab:setting_table}
\end{table}
%\subsection{Influence of thresholding parameter}
%The thresholding parameter \(\Gamma\) determines how far away reads can be from each other and be considered positives. This has some effects on both the training, and the resulting embeddings. Thinking of the embedding as a thread in \(d-\)dimensions, this will act to shorten and thicken the thread. This is illustrated in Figure xx

\subsection{Gargammel settings}

\begin{figure}[htb]
     \centering
     \includegraphics[width=0.99\linewidth]{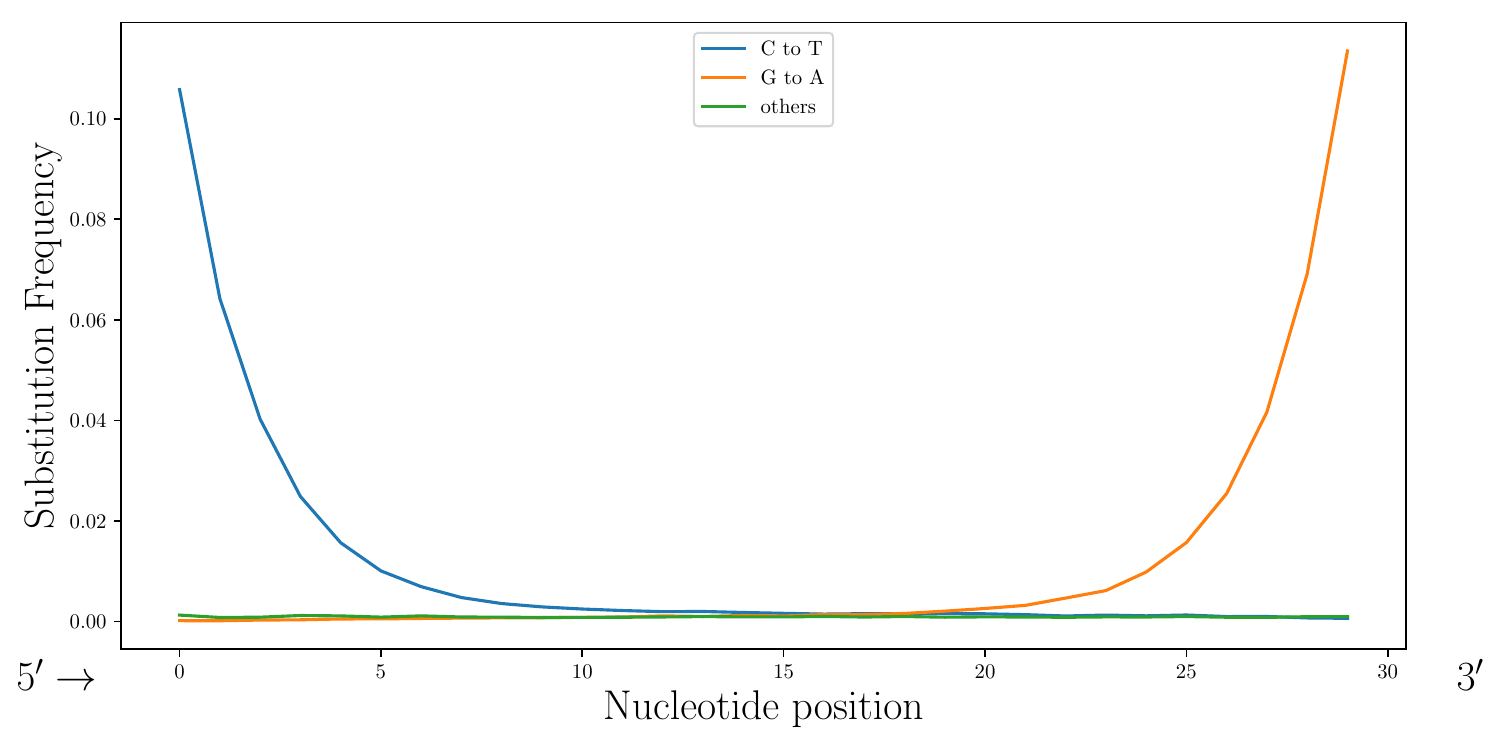}
     \caption{Substitution frequencies of the reads after simulated deamination using the Gargammel read simulator with parameters according to a known damage profile for ancient DNA. }
     \label{fig:sub_freq}
 \end{figure}

We use the Briggs \cite{briggs} damage model. The parameters employed are similar to those cited for  Neanderthal DNA in \cite{briggs}. Our exact choice of parameters to create the aDNA reads with gargammel are listed in Table \ref{tab:gargammel}.
This is done to try to mimic realistic damage patterns in the simulated reads using in performance evaluation.
The deamination pattern in the reads simulated with gargammel is shown in Figure \ref{fig:sub_freq}, where we show the frequency for the substitutions $C\to T$ in the \(5'\) direction and $G\to A$ in the \(3'\)  direction. Comparing this to our aDNA augmentation, where we for the first and last 10 bp increase the substitution rate to 10\%, we can conclude that we are using relatively heavy augmentation. There is an inherent point in not simulating reads based on a model identical to the one used for augmentation to show that we can generalize beyond a highly specific noise pattern.

\begin{table}[htb]
    \centering
    \begin{tabular}{ll}
    \hline
        \textbf{Read argument} & \textbf{Value}   \\
        \hline
        Fragment length (-l)& 30  \\
        
         Coverage (-c)& 20 \\
         \hline
         \textbf{Briggs deamination settings}&   \\
         \hline
         Nick frequency & 0.03   \\
        Length of overhanging ends (geometric parameter)& 0.4  \\
         Probability of deamination of Cs in double-stranded parts&  0.01  \\
         Probability of deamination of Cs in single-stranded parts& 0.7   \\
         \hline
    \end{tabular}
    \caption{Parameter setting used for generating the aDNA reads with Gargammel.}
    \label{tab:gargammel}
\end{table}

\subsection{Computational resources} \label{sec:comp}
The embeddings used in mapping reads from the full \textit{E.\ coli} genome used 16 A100s 40GB, and was trained for 150\,000 iterations, taking approximately 12 hours, and a total of 192 GPUh. This is expensive for a small genome, but continued efforts to improve architecture and optimization strategy could give more reasonable timings. The models trained on 10\% used two A100s, and were trained for 50\,000 iterations, resulting in ~4 GPUh. The models were trained without early stopping, and the numbers may thus be a bit larger than they need to be.
Training the position prediction task is comparably cheap, being done on one A100, for 10\,000 iterations. The most expensive model, the GPT head trained on CReadNet-B, takes 13 minutes in total, and the cheapest, regression on CReadNet-T takes 2.6 minutes. 
The hardware used for inference with the deep learning models is a single A100 Nvidia GPU. The other aligners (BWA-aln, BWA-mem, bowtie2) are CPU-based implementations, and we used 16 cores of an AMD EPYC 7742 CPU, which is the default number of CPU given per GPU on the cluster used.

The embedding plots in Figures \ref{fig:emb} and \ref{fig:noise} were trained in the same way as the read mapping models for the 10\% case. The experiment shown in Figure \ref{fig:2slice} was run on an A2000 laptop GPU with 4GB VRAM, for 3000 iterations.

\subsection{Software}
The code developed for this project was written in Python, using TensorFlow version 2.15 \cite{tensorflow2015}. We used BWA version 0.7.18-r1243-dirty \cite{BWA} and Bowtie2 version 2.5.4 \cite{bowtie}.

In read mapping using Bowtie2, we used the options  \texttt{-p 16 --very-sensitive}, and for BWA-mem, we used \texttt{-k 19 -r 2.5 -t 16}, as recommended in the literature. 
\subsection{CReadNet Architecture}
The description of the encoder in the main text is a bit brief, and in Table \ref{tab:enc} we show the full encoder.

\begin{table}[htb]
    \centering
    \begin{tabular}{lll}
        \hline
        \textbf{Layer} & & \textbf{Output Shape} \\
        \hline
        Input && (batch, 30, 4) \\
        Convolution + LN & \(S = 1, K = 3, C= 64,\) GeLU& (batch, 30, 64) \\
        Stage 1 ResBlocks &\(B = 3, C = 64\) & (batch, 30, 64) \\
        AveragePooling  &\(S= 2, K=2\)& (batch, 15, 64 ) \\
        Convolution & \(K = 3, C= 128\), GeLU& (batch, 15, 128) \\
        Stage 2 ResBlocks &\(B = 3, C = 128\) & (batch, 15, 128) \\
        AveragePooling  &\(S= 2, K=2\)& (batch, 8, 128 ) \\
        Convolution & \(K = 3, C= 256\), GeLU& (batch, 8, 256) \\
        Stage 3 ResBlocks &\(B = 9, C = 256\) & (batch, 8, 128) \\
        AveragePooling  &\(S= 2, K=2\)& (batch, 4, 256 ) \\
        Convolution & \(K = 3, C= 512\), GeLU& (batch, 4, 512) \\
        Stage 3 ResBlocks &\(B = 3, C = 512\) & (batch, 15, 512) \\
        GlobalAveragePooling & &(batch, 4, 512) \\
        Representation (dense) &units= 512, L2 norm, GeLU& (batch, 512) \\
        Embedding (dense) &units = 256, L2 Norm &(batch, 256) \\
        \hline
    \end{tabular}
    \caption{ Architecture of CReadNet-T. Here, $K$ denotes kernel size, $C$ number of convolution filters, $B$ the number of residual blocks in each stage. To scale the model, the number of blocks in stage 3 is increased to 27 (giving CReadNet-S), and the base filter count is increased from 64 to 96 (CReadNet-B). }
    \label{tab:enc}
\end{table}

\subsection{GPT-bit prediction architecture}

This section provides a more detailed description of the GPT model architecture than what is presented in the main paper.
The GPT model takes as input the 512-dimensional output \(Y\) of the three-layer MLP. 
 This output is split into 8 tokens of size 64, denoted as
\[
Y = [Y_1, \dots, Y_8], \quad Y_i \in \mathbb{R}^{64}.
\]
The model also receives a sequence of targets representing the bit-wise representation of the coordinates
\[
T = [t_1, t_2, \dots, t_{N_b}],
\]

and takes all but the last bit \(T_- = [t_1, t_2, \dots t_{N_b-1}]\), and tries to sequentially predict the next bit at each step.
%We use a token size of 64, and split the MLP output into 8 chunks of 64, \(Y_{MLP} = [Y_1, \dots, Y_8]\)
For the bit tokens, we use a trainable embedding \(E \in \mathbb{R}^{2\times64}\), and append it to the MLP tokens, and apply a trainable positional encoding \(E_{pos}\)  to get the combined input 
\[
y = [Y_1, \dots, Y_8,\, t_1E,\, t_2E,\, \dots,\, t_{N_b-1}E] + E_{\text{pos}}.
\]
This input is processed by a Transformer decoder block, with multi-head self-attention (MSA), layer normalization (LN), and a 2-layer feed forward network (FF), finishing with a dense layer with \(b\) units as:
\begin{align*}
a &= \text{LN}(\text{MSA}(y) + y), \\
y &= \text{LN}(\text{FF}(a) + a), \\
Z &= \text{softmax}(\text{Dense}(y)),
\end{align*}

The MSA has a \textit{causal mask} that prevents backward information flow across the bit-tokens, but allows full attention over the MLP tokens.
The output Z has dimensions \(\mathbb{R}^{(8+N_b-1,b)}\), and during training, the first 7 tokens are discarded, and the remaining ones are trained to predict the true bit representation \(T\) of the coordinates by minimizing the categorical cross entropy loss.
The FF network inside the GPT model has two layers, with 256 and 64 units, respectively, with GeLU activation between them. As stated in the main text, we use 2 heads in the MSA, and the final dense layer before the softmax has \(b\) units, bringing the output dimension to \(\mathbb{R}^{(8+N_b,b)}\), in this case. 
Thus, the output is the shifted sequence \(Z = [\tilde{Y}_2, \dots \tilde{Y}_8,\tilde{t}_1, \dots \tilde{t}_{N_b} ]\), and we can extract the predicted bit sequence \([\tilde{t}_1, \dots \tilde{t}_{N_b}]\) to use as the final prediction of the bit-wise representation.

\iffalse
\begin{table}[htb]
    \centering
    \begin{tabular}{lll}
        \hline
        \textbf{Layer} &test  & \textbf{Output Shape} \\
        \hline
        Input && (batch, 512) \\
        LayerNorm +Dense + SilU&& (batch, 2048) \\
        LayerNorm +Dense + SilU&& (batch, 2048) \\
        LayerNorm +Dense + SilU&& (batch, 2048) \\
         MLP output \(Y\) (Dense)&512 units& (batch, 512)  \\

        MLP tokens & Split \(Y \to [Y_1, Y_2, \dots, Y_8]\) & (batch,8,64) \\
        Bit tokens & E()

        %\multirow{3}{*}{$\left\{\begin{array}{c} \\ \\ \end{array}\right.$} & Row 1 \\
        %                         & Row 2 \\
        %                         & Row 3 \\
        %\hline
    \end{tabular}
    \caption{ Architecture of CReadNet-T. Here, $K$ denotes kernel size, $C$ number of convolution filters, $B$ the number of residual blocks in each stage. To scale the model, the number of blocks in stage 3 is increased to 27 (giving CReadNet-S), and the base filter count is increased from 64 to 96 (CReadNet-B). }
    \label{tab:enc}
\end{table}
\fi
%\section{Ethics and Broader impact}

%%%%%%%%%%%%%%%%%%%%%%%%%%%%%%%%%%%%%%%%%%%%%%%%%%%%%%%%%%%%

\end{document}